\documentclass[runningheads]{llncs}

\usepackage{eccv}

\usepackage{eccvabbrv}
\usepackage{bbding}
\usepackage{graphicx}
\usepackage{booktabs}
\usepackage{wrapfig}
\usepackage[linesnumbered,ruled,vlined]{algorithm2e}
\usepackage[accsupp]{axessibility}  
\usepackage[pagebackref,breaklinks,colorlinks,citecolor=eccvblue]{hyperref}
\usepackage{diagbox}
\usepackage{array}
\usepackage{xcolor}
\usepackage{multirow}
\usepackage{caption}

\usepackage{orcidlink}
\DeclareCaptionLabelFormat{andtable}{#1~#2  \&  \tablename~\thetable}

\begin{document}

\title{Out-of-Bounding-Box Triggers: A Stealthy Approach to Cheat Object Detectors} 

\titlerunning{Out-of-Bounding-Box Triggers}

\author{Tao Lin\inst{1,2} \and
Lijia Yu\inst{1} \and
Gaojie Jin\inst{1,2} \and
Renjue Li\inst{1,2} \and
Peng Wu\inst{1,2,}\textsuperscript{\Envelope} \and
Lijun Zhang\inst{1,2,3,}\textsuperscript{\Envelope}}

\authorrunning{T. Lin \emph{et al.}}
\institute{Key Laboratory of System Software (Chinese Academy of Sciences) and State Key Laboratory of Computer Science, Institute of Software, Chinese Academy of Sciences, Beijing, China \and
University of Chinese Academy of Sciences, Beijing, China \and
Automotive Software Innovation Center, Chongqing, China\\
\email{\{lintao,yulijia,gaojie,lirj19,wp,zhanglj\}@ios.ac.cn}}

\maketitle

\begin{abstract}
  
    In recent years, the study of adversarial robustness in object detection systems, particularly those based on deep neural networks (DNNs), has become a pivotal area of research. 
    Traditional physical attacks targeting object detectors, such as adversarial patches and texture manipulations, directly manipulate the surface of the object. 
    While these methods are effective, their overt manipulation of objects may draw attention in real-world applications. 
    To address this, this paper introduces a more subtle approach: an inconspicuous adversarial trigger that operates outside the bounding boxes, rendering the object undetectable to the model. 
    We further enhance this approach by proposing the {\bf Feature Guidance (FG)} technique and the {\bf Universal Auto-PGD (UAPGD)} optimization strategy for crafting high-quality triggers. 
    The effectiveness of our method is validated through extensive empirical testing, demonstrating its high performance in both digital and physical environments. 
    The code and video will be available at: \url{https://github.com/linToTao/Out-of-bbox-attack}.
  \keywords{Universal physical adversarial attack \and Out-of-bounding-box trigger\and Object detection}
\end{abstract}

\section{Introduction}
\label{sec:intro}
Object detection aims to identify instances of semantic objects typically within images or video clips. It finds widespread applications across a range of domains, including face recognition {~\cite{deng2019retinaface,schroff2015facenet,wang2018cosface,deng2019arcface,huang2023safari}}, object tracking {~\cite{li2019siamrpn++,li2018high,wang2019fast}}, and autonomous driving {\cite{chen2015deepdriving,grigorescu2020survey,zhang2023trajpac}}. 
Especially in autonomous driving systems, object detectors play a crucial role in tasks such as recognizing traffic signs, pedestrians, vehicles, and traffic lights. 
Nonetheless, in recent years, security concerns have emerged regarding object detectors, primarily due to the susceptibility of deep neural networks (DNNs) to adversarial examples (AEs). 
AEs are intentionally crafted inputs designed to deceive DNNs into making incorrect predictions. 

\begin{figure}[t]
   \centering
   \includegraphics[width=0.95\linewidth]{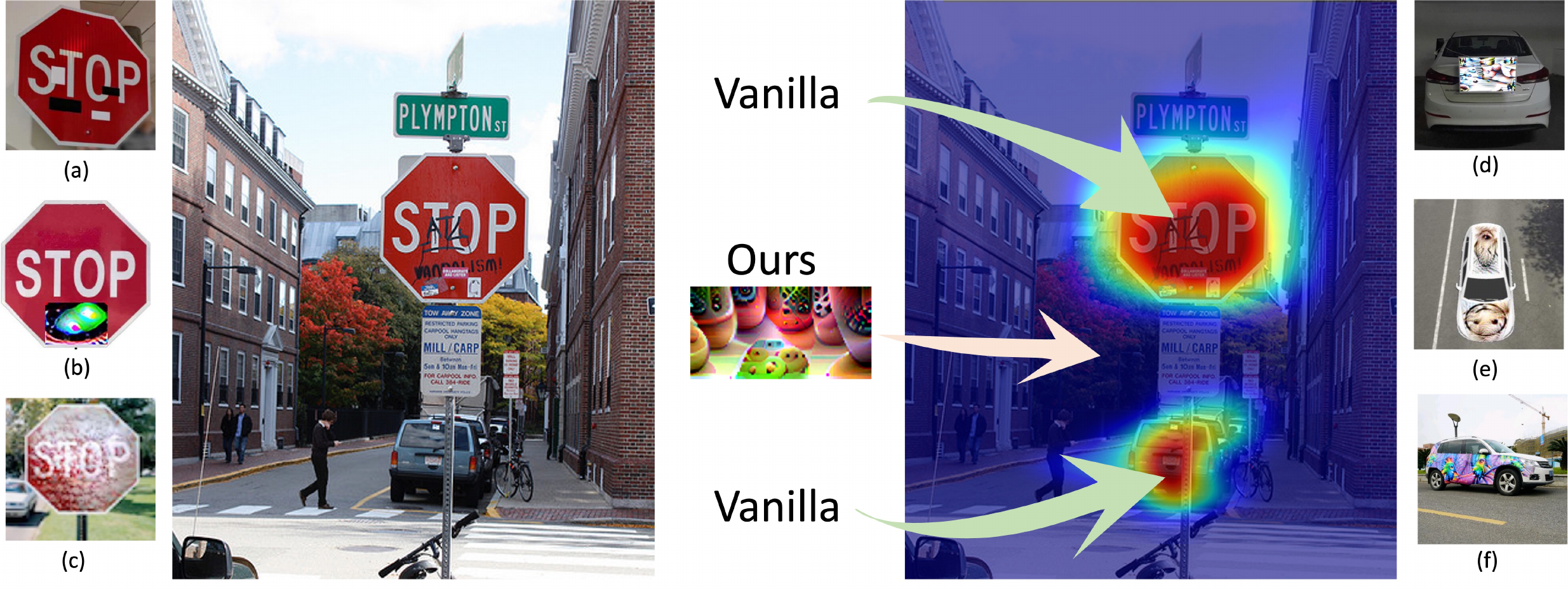}
   \caption{ An illustration of vanilla object attack methods and our novel approach. Typically, vanilla attack methods target the most informative regions of an object, such as the central, highly visible areas of the target object, e.g., the stop sign in (a)~\cite{eykholt2018robust}, (b)~\cite{zhao2019seeing}, (c)~\cite{duan2020adversarial}, and the car in (d)~\cite{hoory2020dynamic}, (e)~\cite{wang2021dual}, (f)~\cite{huang2020universal}.  
   In contrast, our method adopts a more covert strategy, focusing on attacking the peripheral edge areas surrounding the stop sign.}
   \label{intro}
   
\end{figure}

Early research mainly focused on studying AEs like adversarial patch {~\cite{eykholt2018robust,liu2019perceptual,hu2021naturalistic,zolfi2021translucent,lei2022using}}, optical {~\cite{zhong2022shadows,lovisotto2021slap}} and texture {~\cite{duan2020adversarial,hu2022adversarial}}, which are targeted at the objects. These AEs involve the computation of perturbations applied directly to the objects to mislead a DNN-based detector.
However, a direct attack on the object as shown in \figureautorefname~\ref{intro}
, along with its subsequent attempt to deceive the detector, is overly conspicuous and cannot be deployed without raising suspicion in real-world practical scenarios.

In this work, we provide an alternative perspective: \emph{is it possible to introduce an adversarial trigger in an inconspicuous area outside the object while still successfully deceiving the detector?}

Actually, crafting an adversarial trigger within the areas beyond the object boundaries is even more demanding.
On one hand, for dense prediction networks like object detectors, the output results are more reliant on the red regions of the heatmap in \figureautorefname~\ref{intro}.
This implies that tailoring the adversarial trigger by utilizing irrelevant information, like the blue regions of the heatmap, is sharply challenging.
More difficultly, our trigger is designed to conform to practical size and shape constraints. 
Nevertheless, for exactly the same reasons, our attack method gets more delusive and harder to defense.


This paper is focused on the development of robust and inconspicuous triggers to attack typical 
object detectors deployed in practical scenarios. 
To tackle the challenge mentioned above, we present a novel technique called Feature Guidance (FG).
In this approach, we guide the feature layers with an adversarial trigger to resemble those with the object hidden. 
And we propose a optimization strategy, called Universal Auto-PGD (UAPGD), applicable to universal attacks. This optimization strategy allows for the selection of an appropriate timing to halve the step size, based on the oscillation of the loss function.

To assess the effectiveness of our approach in both digital and physical domains, we trained the adversarial trigger and evaluated the corresponding metrics on COCO {\cite{lin2014microsoft}} dataset and Carla {~\cite{dosovitskiy2017carla}}, respectively.  
We also evaluated the robustness of the trigger by recording videos using an onboard camera in the real world.

Overall, this paper makes the following contributions:
\begin{itemize}
    \setlength{\itemsep}{0pt}
    \setlength{\parsep}{0pt}
    \setlength{\parskip}{0pt}
    \item This work studies the realm of adversarial robustness in object detection, providing a novel perspective by exploring the concept of an adversarial trigger applied outside an object.
    \item 
    We propose a novel attack method designed to generate triggers that achieve excellent continuous attack effects and demonstrate high robustness in real world. The key enabler of our method is the utilization of similarity between feature layers and adaptive step size decay to endow the trigger with the ability to mislead the detector.
    \item An extensive set of empirical results are provided to demonstrate that our attack exhibits strong performance and robustness in both the COCO dataset and Carla simulator, as well as in real-world scenarios.
\end{itemize}

\section{Related work}
Adversarial examples (AEs), first discovered by Szegedy \emph{et al.}{~\cite{szegedy2013intriguing}}, involve carefully crafted inputs inducing model misclassifications. 
Transferability{~\cite{xie2019improving,wang2021enhancing,jia2022adv,zhang2022improving}} of AEs, allows them to deceive networks with different hyperparameters or architectures. 


{\bf Digital adversarial examples} are generated by directly manipulating any pixels in digital images prior to feeding them into the model. 
Adversarial attacks can be classified into black-box attacks and white-box attacks based on whether the internal structure and parameters of the target model are known. White-box attacks {~\cite{carlini2017towards,goodfellow2014explaining,madry2017towards,moosavi2016deepfool,dong2018boosting,kurakin2016adversarial,jin2023randomized,jin2022enhancing}} primarily rely on the computation of model gradients. Black-box methods, on the other hand, depend on querying the unknown model by observing the prediction {~\cite{dong2018boosting,dong2019evading,xie2019improving,wu2020skip,wu2020boosting,pomponi2022pixle,zhu2023efficient,liang2022parallel}} or leveraging an agent model to imitate the threaten model {~\cite{chen2017zoo,brendel2017decision,ilyas2018black,chen2020hopskipjumpattack,zhang2022towards,kahla2022label}}. Recently, Croce \emph{et al.} {\cite{croce2020reliable}} proposed AutoAttack, a parameter-free method that combines novel white-box and black-box attack strategies.
The works of applying digital adversarial examples to object detectors have been explored in ~\cite{im2022adversarial} and ~\cite{shapira2023phantom} that introduce perturbations to input images to mislead detectors. ~\cite{shi2023reinforcement} generate more subtle perturbations via reinforcement learning, successfully attacking the detector.

{\bf Physical adversarial examples} are crafted to be printed out and captured by sensory equipment like camera {~\cite{kurakin2016adversarial,li2019adversarial}} and LiDAR {~\cite{cao2019adversarial,tu2020physically,liu2023slowlidar}}. These examples pose a real-world threat due to their ability to adapt to camera and scenario variations. Early studies focused on attacking recognition models {~\cite{sharif2016accessorize,eykholt2018robust}}. Recent research has explored physical attacks in the more challenging context of object detection. Perturbations in this domain include adversarial patches {~\cite{thys2019fooling,xu2020adversarial,hu2021naturalistic,zhao2019seeing,aich2023leveraging}}, adversarial camouflage textures {~\cite{huang2020universal,wang2021dual,wang2022fca,hu2022adversarial}}, and optical attacks {~\cite{lovisotto2021slap,zhu2021fooling,li2023physical,wang2023rfla}}. To improve robustness against various transformations in physical settings, Athalye \emph{et al.} {~\cite{athalye2018synthesizing}} introduced Expectation Over Transformation (EOT), widely used in physical attacks, to enhance the effectiveness of adversarial attacks. 

Prior physical attack methods need to be applied directly to the objects so that these methods are expected to yield favorable results but they are prone to raising suspicion in practical scenarios. Unlike these works, our method constructs a universal attack trigger that can deceive the detector disregarding the specific object without altering the appearance of itself, and can be discreetly deployed in the real world.

The works that exhibit the closest alignment with our motivation are Dpatch ~\cite{liu2018dpatch} and ~\cite{lee2019physical} (after this referred to as Dpatch2).
However, we must emphasize that our work is independent of them,  highlighting the difference between our work and theirs in Appendix F. Furthermore, in the experiments section, we compare the attack effectiveness of our method with Dpatch2 ~\cite{lee2019physical} on the COCO dataset.

\section{Methodology}
\label{method}

\begin{figure}[!t]
    
    \centering

    \begin{subfigure}{0.49\linewidth}
       \centering
       \includegraphics[width=1\linewidth]{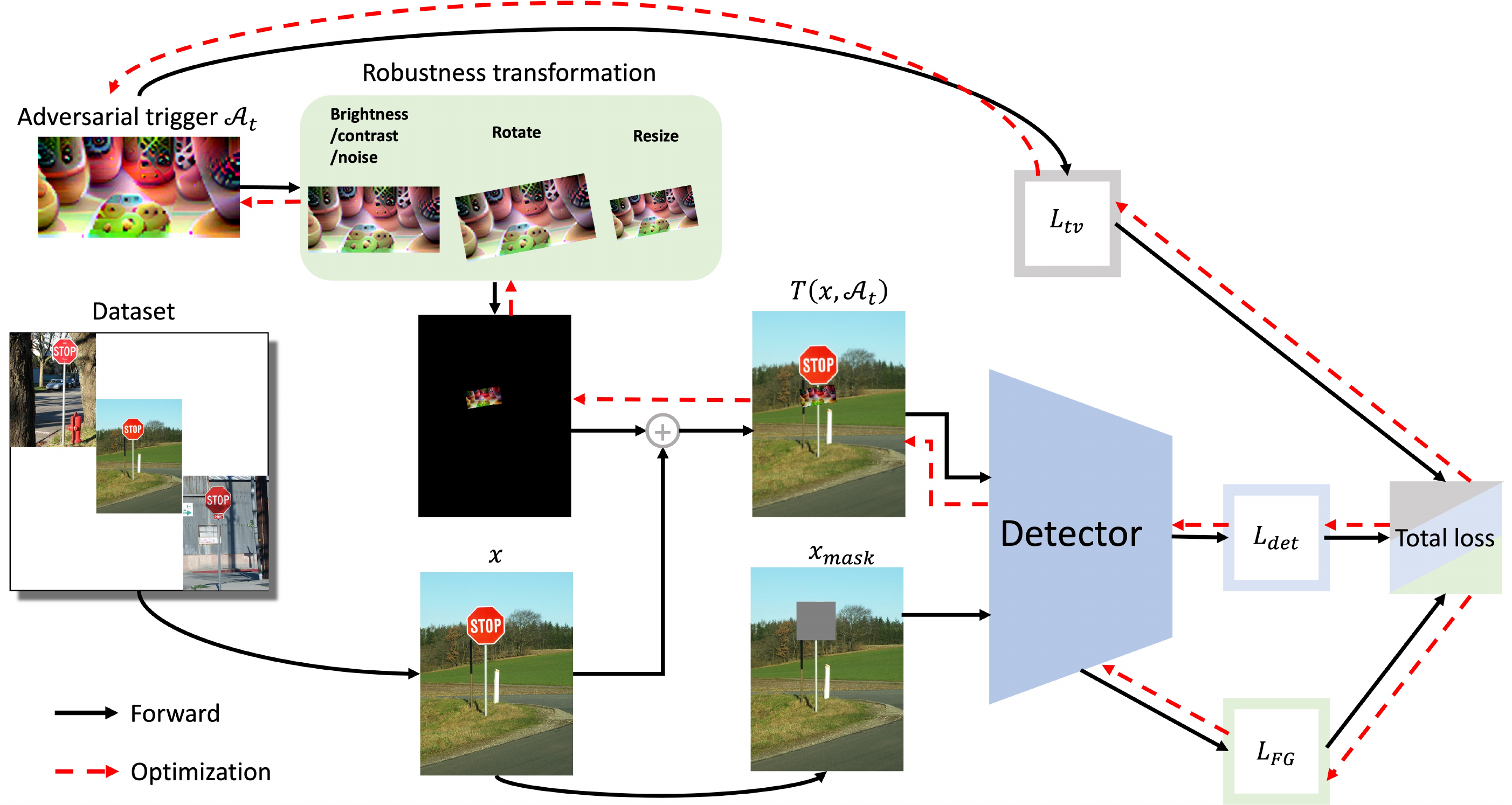}
       \caption{}
       \label{framework}
   \end{subfigure}
    \centering
   \begin{subfigure}{0.49\linewidth}
       \centering
       \includegraphics[width=0.9\linewidth]{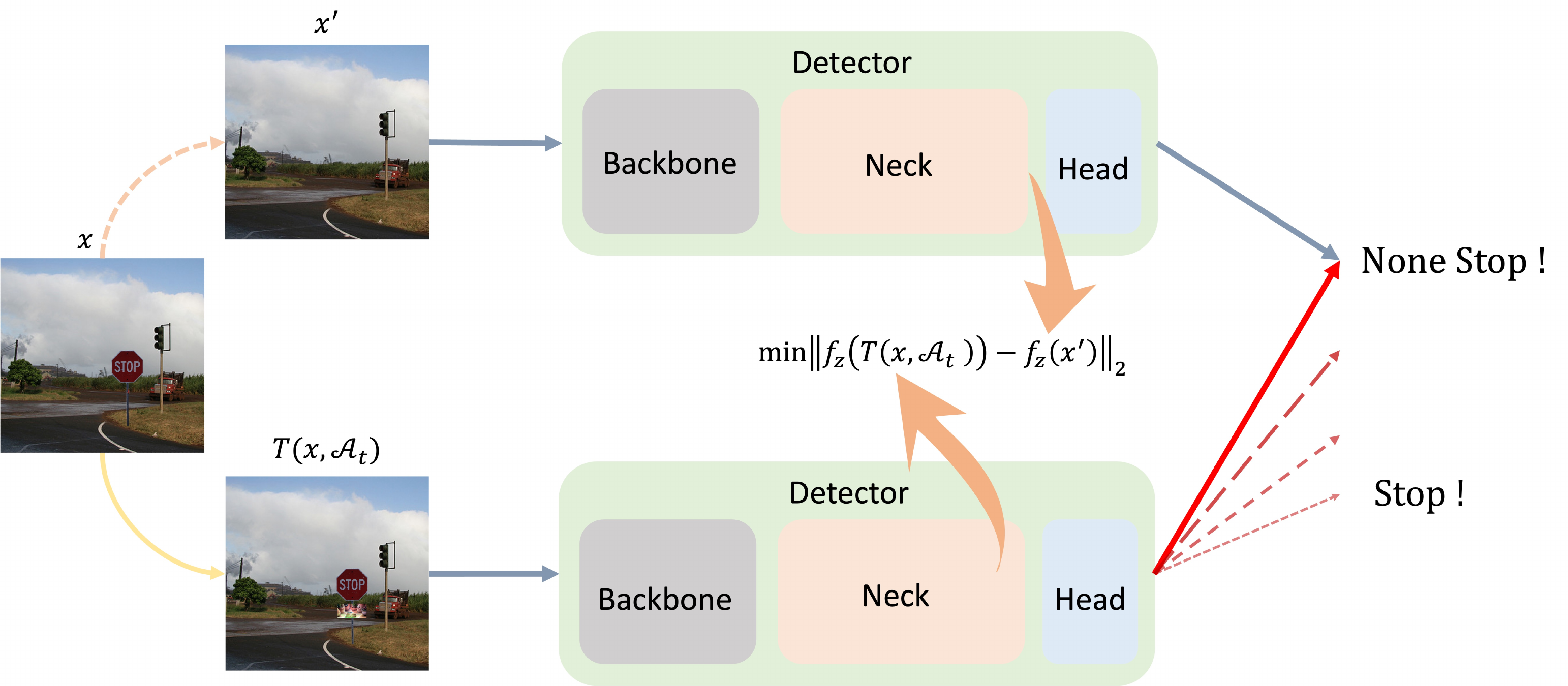}
       \caption{}
       \label{FG_motivation}
   \end{subfigure}
    \caption{(a) shows the overview of our adversarial trigger generation framework. 
    We deploy the adversarial trigger onto each image in the dataset with robustness transformation. 
    With a fixed detector, our method optimizes the adversarial trigger through the obtained adversarial gradient. 
    (b) illustrates the motivation and calculation method of Feature Guidance. 
    Taking a stop sign as an example, for a given image $x$, 
    minimizing $\|f_{z}(T(x,\mathcal{A}_{t}))-f_{z}(x')\|_2$ can mislead a detector to classify the stop sign image $T(x,\mathcal{A}_{t})$ as {\bf none stop}. 
    }
    \label{All_framework}

\end{figure}

In this work, we aim to generate physical adversarial triggers that can be naturally deployed in an inconsequential position. 
These adversarial triggers are designed not to destroy the object information but still deceive the model effectively.
In this section, we commence by presenting the problem formulation in Section 3.1. 
Subsequently, \figurename~\ref{framework} provides a comprehensive illustration of our framework. 
Our primary focus is on the white-box attack paradigm, which necessitates an intricate optimization process. 
Given an object detection model, we design a loss function to determine the gradient direction for adversarial attacks in Section 3.2. 
This loss function is then utilized iteratively to update the adversarial trigger. 
In the pursuit of strengthening attack performance and enhancing the continuity of attacks in video-based scenarios, we introduce a novel method named ``Feature Guidance" (FG) and an additional loss term $L_{FG}$ in Section 3.2. 
To bolster the overall robustness and deployability of our approach, we also employ two indispensable techniques, EOT and total variation loss in Section 3.2. 
Furthermore, our methodology draws inspiration from the work of \cite{croce2020reliable} in their Auto PGD framework during the training process,  we thus modify the conditions of step size decay based on the universal training mode (Section 3.3).

\subsection{Problem Formulation}

{Adversarial attacks on object detection can have diverse objectives, encompassing but not limited to inducing misalignment in bounding boxes, reducing classification confidence, or provoking misclassification.}
{Our goal is to generate adversarial examples for a specific class, with the aim of causing the object detection algorithm to fail in identifying objects belonging to that class.}

Given an input image $x\in\mathbb{R}^d$ with the ground truth $y=\{(y_i^{\rm{coor}},y_i^{\rm{cls}})\}_{i\in[n]}\subset\mathbb{R}_+^4\times\mathbb{Z}_+$, $n$ is the number of objects in the image, $y_i^{\rm{coor}}$ contains top-left and bottom-right coordinates of the bounding box in the image, and $y_i^{\rm{cls}}$ is the class of object.
Then, the DNN-based object detector can be represented as the following form:  $$f:\mathbb{R}^d\to 2^{\mathbb{R}_+^4\times\mathbb{Z}_+\times[0,1]^2}$$
For a given image $x$, the output of $f$ can be represented as
$\Tilde{y}=\{(\Tilde{y}_i^{\rm{coor}},\Tilde{y}_i^{\rm{cls}},\Tilde{y}_i^{\rm{conf}})$ $\}_{i\in [n]}\subset\mathbb{R}_+^4\times\mathbb{Z}_+\times[0,1]^2$
, where $\Tilde{y}_i^{\rm{conf}}$ denotes the confidence associated with the predicted box coordinates $\Tilde{y}_i^{\rm{coor}}$ and the predicted class $\Tilde{y}_i^{\rm{cls}}$ for $i$-th object.
We aim to generate a universal adversarial trigger $\mathcal{A}_{t}\in[0,1]^{3\times U\times V}$, where $U, V\in \mathbb{Z_+}$ are the given dimensions, for the given specific class $t$, such that these adversarial triggers can cause $f$ to fail in recognizing objects of class $t$. 
Then, the problem can be formulated as an optimization task, represented as follows:
\begin{equation}
\label{real_opt}
    argmin_{\mathcal{A}_{t}} \mathbb{E}_{x\sim D}[|f(T(x,\mathcal{A}_{t}))\cap(\mathbb{R}_+^4\times\{t\}\times[0,1]^2)|]   
\end{equation}
where $|A|$ denotes the cardinality of the set $A$, $D$ represents the input distribution, and $T$ (which will be elaborated in the following) is a transform operation to incorporate the adversarial trigger into the original image. 
{In contrast to prior studies based on adversarial patches, our approach avoids placing the adversarial trigger on the informative surface, e.g., traffic signs, thereby preventing any unnecessary attention.
We would like to point out that this kind of attack method is challenging because the adversarial trigger can only be stuck on the unimportant places of the image.}

{\bf Transform operation.} 
We utilize a differentiable operation $T$ to render the adversarial trigger $\mathcal{A}_{t}$ onto an image $x$. 
Let the object with class $t$ be located in the box $B$ of the image, we use the following two-step to add the adversarial trigger to the image:\\ 
(1) We can select an appropriate area $C(B)$ in the image where $C$ is a given function, such position does not overlap with the surface of the object; \\
(2) We cover the box $C(B)$ part of the image $x$ with adversarial trigger $\mathcal{A}_{t}$, but considering that the adversarial trigger and the box are not the same size, we need to apply the following affine matrix to $\mathcal{A}_{t}$:
%
%
\begin{equation}
    \theta=\left(\begin{array}{ccc}
       S_{h}cos\alpha  & -S_{h}sin\alpha & T_{h} \\
       S_{v}sin\alpha  & S_{v}cos\alpha & T_{v} \\
       0 & 0 & 1
    \end{array}
    \right)
    \label{matrix}
\end{equation}
$S_{h}$ and $S_{v}$ denotes the scaling factors utilized to resize the adversarial trigger, $\alpha$ represents the angle of rotation, $T_{h}$ and $T_{v}$ are the distance adversarial trigger need to shift. These parameters are all computed by $C(B)$. The more specific remark regarding $C(B)$ will be detailed in Appendix B.



Note that our attack method has two characteristics: Our attack is out of the object; our attack area changes depending on the position of the object. We were the first to attack target detection with such a method.

\subsection{Adversarial gradient with Feature Guidance}
\label{sec1}

In this paper, we focus on utilizing white-box attack methods to search for adversarial gradients and optimize the adversarial trigger. 
Additionally, we provide black-box empirical results in Appendix D that the trigger is generated by a surrogate model. 
As illustrated above, we render the adversarial trigger onto images and feed them to the detectors, take the derivative of the loss function, and use it to update the adversarial trigger. 
The loss function consists of the following parts:

{\bf Adversarial loss of detector output.}  
As mentioned above, the detector outputs the predicted class and bounding boxes and their confidence. Considering most of the detectors will not output the predicted class and boxes with a confidence lower than a threshold, we want images with the adversarial trigger to make the confidence as low as possible.


Let $f^{\rm{conf}}_{i,coor}(x)$ and $f^{\rm{conf}}_{i,cls}(x)$ is the confidence of the $i$-th object on image $x$,   we can define the first part of our loss function as follows:
\begin{equation}
L_{det}(\mathcal{A}_{t})=\mathbb{E}_{x\sim D}[\max_{i:\Tilde{y}_i^{cls}=t}f^{\rm{conf}}_{i,coor}(T(x,\mathcal{A}_{t}))f^{\rm{conf}}_{i,cls}(T(x,\mathcal{A}_{t}))]
\end{equation}

{\bf Feature guidance.} 
The biggest difficulty with an attack outside the object is that the features of the original object on the image will not be destroyed, so the network will still notice the presence of the object.


We gain the following insights from the network structure to address this issue. A DNN-based detector is usually designed with three modules: Backbone, Neck, and Head. The Backbone module extracts global features from an input image. These features are then fused through the Neck module to produce multiple feature maps with varying receptive fields. The Head module subsequently makes predictions based on the feature maps. This modular structure imitates how people often perceive a comprehensive horizon (like the Backbone module) and focus on varying attractive areas (like the Neck module). Subsequently, the attractive areas will contribute to making perceptions (like the Head module). Hence, in order to provoke model errors, it is imperative to deliberately induce information misleading prior to inputting the head module.

To achieve that, we use the following method, called {\bf Feature guidance}, as shown in \figurename~\ref{FG_motivation}. For a given image $x$ with a target object $t$, we can get the image $x'$, which differs from $x$ only with the object $t$ missing.
Then, if we can elaborate our adversarial trigger $\mathcal{A}_{t}$ to guide the feature map $f_z(T(x,\mathcal{A}_{t}))$ 
as close to $f_z(x')$ generated by the Neck module of the detector, it will have a desirable impact on the Head module such that the stop sign cannot be detected. 

As illustrated in \figureautorefname~\ref{framework}, for the sake of convenience, we derive $x_{mask}$ that exhibits an analogous effect to $x'$ simply by covering the bounding box position in the ground truth of $x$ with a gray rectangle.

In order to achieve {Feature guidance}, we need to make $\mathcal{A}_{t}$ to minimum such loss function:
$$L_{FG}(\mathcal{A}_{t})=E_{x\sim D}[\|f_{z}(T(x,\mathcal{A}_{t}))-f_{z}(x_{mask})\|_2]$$

{\bf Other skills} 
To enhance the robustness of our adversarial trigger $\mathcal{A}_{t}$, we incorporate diverse variations, including random noise, contrast, brightness, and rotation, into the generated trigger prior to its application on images, thus simulating real-world conditions. To address printing challenges, we employ TV loss~\cite{mahendran2015understanding} on the adversarial trigger to promote smoothness. TV loss is defined as


\begin{equation}
    L_{tv}(\mathcal{A}_{t})=\sum\limits_{i,j}\sqrt{(\mathcal{A}_{t}^{i,j}-\mathcal{A}_{t}^{i+1,j})^2+(\mathcal{A}_{t}^{i,j}-\mathcal{A}_{t}^{i,j+1})^2}
\end{equation}
where the subindices $i$ and $j$ represent the pixel coordinate of the trigger $\mathcal{A}_{t}$

{\bf Overall,} we will get the adversarial trigger $\mathcal{A}_{t}$ by optimizing following problem:
{\small
\begin{equation}
\label{youhua}
   \min_{\mathcal{A}_{t}\in[0,1]^{3\times U\times V}}L_{det}(\mathcal{A}_{t})+\lambda_{FG}L_{FG}(\mathcal{A}_{t})+\lambda_{tv}L_{tv}(\mathcal{A}_{t})
\end{equation}
}
where $\lambda_{FG}$ and $\lambda_{tv}$ are hyperparameter.

\begin{algorithm}[t]
  \SetKwData{Left}{left}\SetKwData{This}{this}\SetKwData{Up}{up}
  \SetKwFunction{Union}{Union}\SetKwFunction{FindCompress}{FindCompress}
  \SetKwInOut{Input}{Input}
  \SetKwInOut{Output}{Output}

  \Input{Threaten network $f$, 
  initial step size $\eta$, 
  the number of epoch $N_{epoch}$, 
  dataset $D=\cup_{j=1}^{k} D_j$ which was divided into k batches, 
  the hyperparameter in UAPGD $l_c,l_o$, 
  initial adversarial trigger $\mathcal{A}^0_{t}$.
  }
  \Output{Adversarial trigger $\mathcal{A}_{best}$}
  \BlankLine
  

  $S_c\leftarrow \phi$, 
  $S_o\leftarrow\phi$, 
  $L_{best}\leftarrow \infty$,   
  $\mathcal{A}_{best}\leftarrow \mathcal{A}^{0}_{t}$
  
 \For{$i\in [N_{epoch}]$}{
 
    $L_{i}\leftarrow 0$
    
    $\mathcal{A}^{i,0}_{t}\leftarrow\mathcal{A}^{i-1}_{t}$
    
    \For{$j\in[k]$}{
        Get $L_{all}(\mathcal{A}^{i,j-1}_{t},D_j)$ via $f$
    
        $L_{i} \leftarrow L_{i} + L_{all}(\mathcal{A}^{i,j-1}_{t},D_j)/k$
        
        $\mathcal{A}^{i,j}_{t} \leftarrow \mathcal{A}^{i,j-1}_{t}-\eta\cdot\textbf{sign}(\frac{\nabla L_{all}(\mathcal{A}^{i,j-1}_{t},D_j)}{\nabla \mathcal{A}^{i,j-1}_{t}})$
        
        $\mathcal{A}^{i,j}_{t} \leftarrow Clamp(\mathcal{A}^{i,j}_{t},0,1)$
        
    }
    $\mathcal{A}^{i}_{t} \leftarrow \mathcal{A}^{i,k}_{t}$


    \If{$L_{i}<L_{best}$}{
        $\mathcal{A}_{best}\leftarrow \mathcal{A}^{i}_{t}$ and $L_{best}\leftarrow L_{i}$
    }
    
    $S_o = S_o\cup L_{i}$ 
    
    \If{$|S_o|=l_o$}{
        $L_{min} \leftarrow \min_{L\in S_o}\{S_o\}$
        
        $V \leftarrow \mathrm{Var}_{L\in S_o} \{S_o\}$

        $S_c = S_c\cup(L_{min},V)$ 
        
        $S_o\leftarrow\phi$

    }
    \If{$|S_c|=l_c$}{
    
        \tcp{Condition 1 $\mathrm{and}$ Condition 2 are introduced in \sectionautorefname~\ref{optpro}}
        
        \If{Condition 1$(S_c)$ $\mathrm{and}$ Condition 2$(S_c)$}{
            $\eta \leftarrow \eta/2$ and $\mathcal{A}^{i}_{t}\leftarrow \mathcal{A}_{best}$
        }
        
        $S_c\leftarrow\phi$ 
    }
        
  }
Return: $\mathcal{A}_{best}$.
  \caption{UAPGD}
  \label{algo_disjdecomp}

\end{algorithm}

\subsection{Universal Auto-PGD}
\label{optpro}

Let $L_{all}=L_{det}+\lambda_{FG}L_{FG}+\lambda_{tv}L_{tv}$, we will use gradient descent to find the adversarial trigger via minimizing the loss function $L_{all}$. 
%





Considering the instability that appeared in the experiment when optimizing multiple terms of $L_{all}$ with conventional methods, we propose the following optimization strategy called Universal Auto-PGD (UAPGD):
%




At the beginning of optimizing, we fix the following values: comparison length $l_c$, observation length $l_o$, the slack variable $\epsilon_1,\epsilon_2$, let the initial step size be $\eta$, and we record the following values: \\
(1): After epoch $j$, record the value of the loss function, name it $L_{j}$;\\
(2): After epoch $i*l_o$ where $i\in Z_+$, record:
\begin{equation*}
\begin{array}{cc}
L_{min,i}&=min_{j\in[(i-1)*l_o+1, i*l_o]}\{L_j\} \\
V_{i}&=\mathrm{Variance} [\{L_j\}_{j\in[(i-1)*l_o+1, i*l_o]}]  
\end{array}
\end{equation*}

In order to prevent violent oscillations around the local optimum during the optimization process, we will halve the step size 
when the following conditions are met:\\
(Con 1): $L_{min,j}\ge L_{min,q}-\epsilon_1$ for all $(j-l_c)<q<j$;\\
(Con 2): $V_j\ge V_{q}-\epsilon_2$ for all $j-l_c<q<j$.\\
Due to the stochasticity from the robustness transformation mentioned in {\bf{Other skills}}, the value of the loss function may fluctuate so that we incorporate the slack variable $\epsilon_1,\epsilon_2$ into numerical comparison.

Intuitively, these two conditions respectively describe the exploration and exploitation aspects in the optimization process. 
Con 1 means that the value of the loss function has not decreased much, indicating the necessity of a transition from exploring the entire feasible space to focusing on local optimization.
Con 2 characterizes the oscillation of the loss function
, of which the amplitude per $l_c$ steps has not decreased much.
This implies the imminent conclusion of the algorithm's exploitation in a local space. The simultaneous occurrence of c1 and c2 means that the adversarial trigger jumps near one of the optimal points, but the step size is too large to fall steadily near the best advantages. Therefore, at this point, we will halve the step size.


Our approach is somewhat similar to AutoAttack \cite{croce2020reliable}, but there are several differences: 
(1): Our method is for universal attack, AutoAttack is for attacking a single image;
(2): We consider the amplitude of the oscillation during the optimization process and use this as an indicator of decline;
(3): We include the slack variables in order to mitigate potential misleading phenomena caused by experimental randomness. For example, the oscillation of the loss function often easily cause $L_j$ to shows a downward trend.


The overall process of our algorithm is shown in \algorithmautorefname~\ref{algo_disjdecomp}.
 And we use $L_{all}(\mathcal{A}_{t},D_j)$ to represent calculating $L_{all}$ on an input batch $D_j$. 
 
\section{Experiment}


In this section, we choose the stop sign as the target object to evaluate the effectiveness of our adversarial attack method. 
The stop sign is a commonly used object in the evaluation of physical adversarial attacks, as it is critical to traffic safety. 
We begin by detailing the implementation settings of our evaluation. 
This is followed by an extensive set of experiments, utilizing the proposed adversarial trigger in the digital space, specifically on the COCO dataset \cite{lin2014microsoft}. 
Furthermore, we extend the application of our method to real-world scenarios, i.e., we record videos to demonstrate that our method can attack the object detector in both simulators and real-world scenes. 
Meanwhile, we evaluate our attack method across varying distances and angles. See Appendix A and B for the basic experimental setting, the results of additional attacks on other objects, and the detail of selecting C(B) to place the trigger on the stop sign.

\subsection{Digital experiment}
\label{mainre}


In this subsection, our experiments are carried out using the COCO dataset {\cite{lin2014microsoft}}. 
From this dataset, we extract a total of 1100 images featuring stop signs. 
Of these, 600 images constitute the training set, which is utilized to generate the adversarial trigger. 
The remaining 500 images are designated as the test set, serving to evaluate the effectiveness of the generated adversarial trigger.

%


To demonstrate the superiority of our algorithm, we adopt the following methods to generate adversarial triggers:  optimizing the adversarial trigger using PGD \cite{madry2017towards} or UAPGD with a minimization of the loss function $L_{det}+\lambda_{tv} L_{tv}$ or $L_{all}$. This included comparisons between our method and the normal PGD under the condition of placing the perturbation outside the object, as well as ablation experiments with varying hyperparameters.
Specifically, for the YOLOv3(YOLOv5) model, we choose $\lambda_{FG}$ values of 0.1, 0.5, and 1.0(0.05, 0.1, and 0.2). 
%
The optimization process utilized the initial step size of 8/255 and 16/255.
We calculate the number of undetected images among all the images as a metric called attack success rate(ASR). 

\begin{figure}[!htbp]

\centering
  \begin{minipage}[t]{0.5\linewidth}
    \centering
    \includegraphics[width=\linewidth]{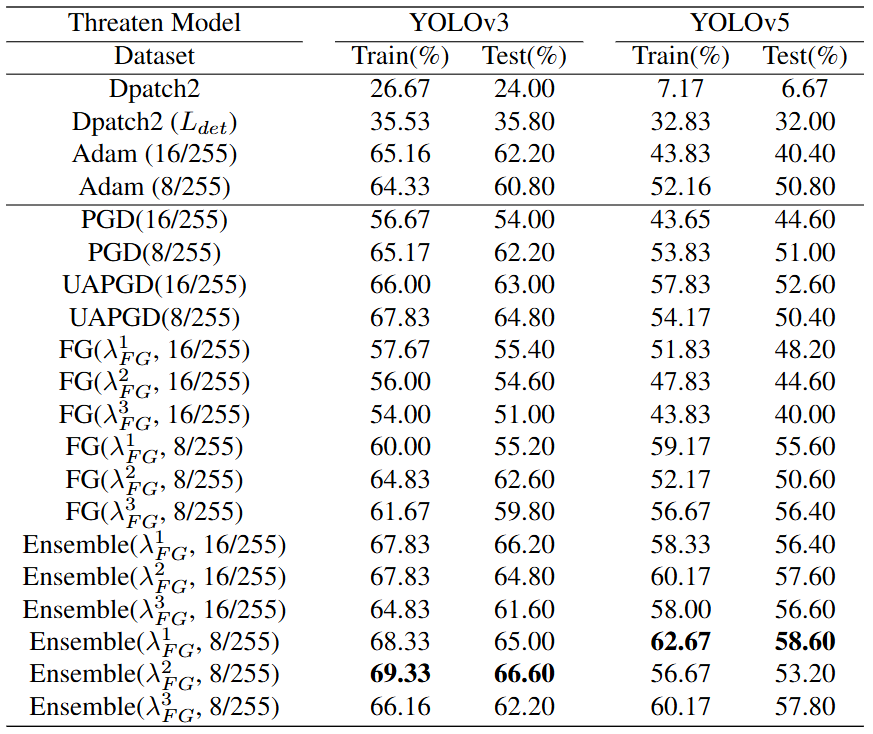}
  \end{minipage}
\raisebox{0.8\height}{
  \begin{minipage}[t]{0.4\linewidth}
      \captionsetup[subfigure]{labelformat=empty}
       \centering
       \begin{subfigure}{0.48\linewidth}
           \centering
           \includegraphics[width=0.8\linewidth]{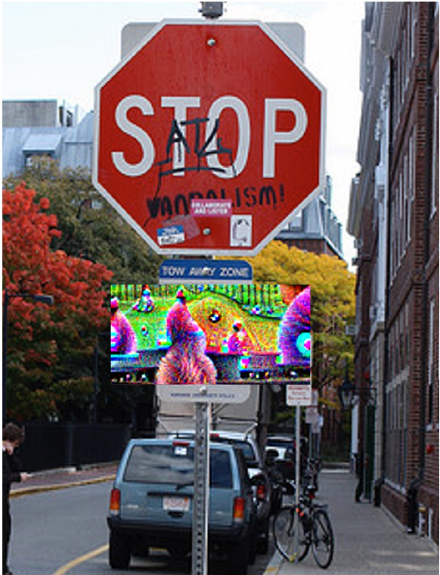}
           \caption{PGD}
           \label{PGD_16}
       \end{subfigure}
       \centering
       \begin{subfigure}{0.48\linewidth}
           \centering
           \includegraphics[width=0.8\linewidth]{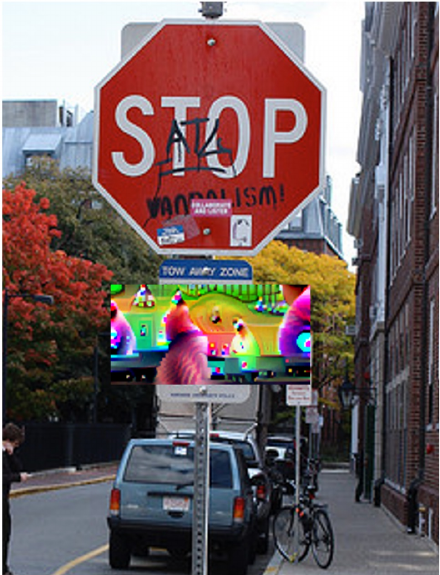}
           \caption{UAPGD}
           \label{UAPGD_16}
       \end{subfigure}
       \centering
       \begin{subfigure}{0.48\linewidth}
           \centering
           \includegraphics[width=0.8\linewidth]{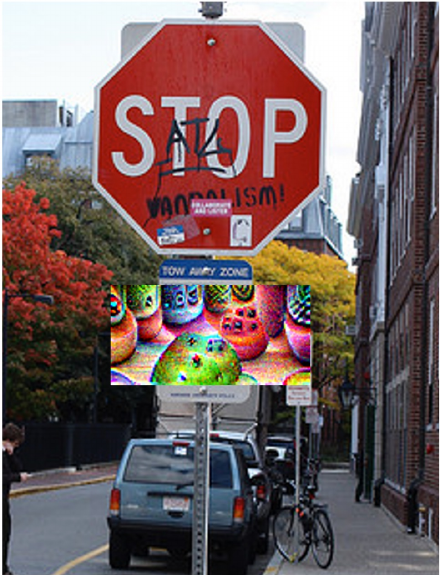}
           \caption{FG}
           \label{FG_16}
       \end{subfigure}
       \centering
       \begin{subfigure}{0.48\linewidth}
           \centering
           \includegraphics[width=0.8\linewidth]{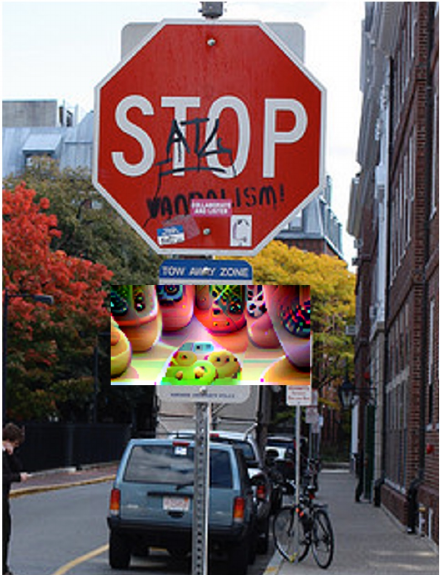}
           \caption{Ensemble}
           \label{all_8}
       \end{subfigure}
  \end{minipage}
}
\caption{The table on the left displays the ASR values of triggers, which are generated by various settings, on the training and testing sets of the COCO dataset. The values in parentheses represent the attack step size, and we employ different $\lambda_{FG}^i$ values for YOLOv3 and YOLOv5. For YOLOv3, we set $\lambda_{FG}^1=0.1,\lambda_{FG}^2 = 0.5,\lambda_{FG}^3=1.0$; for YOLOv5, we set $\lambda_{FG}^1=0.05,\lambda_{FG}^2 = 0.1,\lambda_{FG}^3=0.2$. Ensemble denotes the combined usage of FG and UAPGD. 
Dpatch2 is the original approach from ~\cite{lee2019physical} and Dpatch2($L_{det}$) denotes the original approach adopts the $l_{det}$ as the optimization objective.
The bolded values indicate the optimal results for each respective column in the table. 
There are some results of triggers on the right generated under different settings.}
\label{digital result and trigger}
\end{figure}

\begin{figure}[!t]
   \centering
   \begin{subfigure}{0.22\linewidth}
       \centering
       \includegraphics[width=1\linewidth]{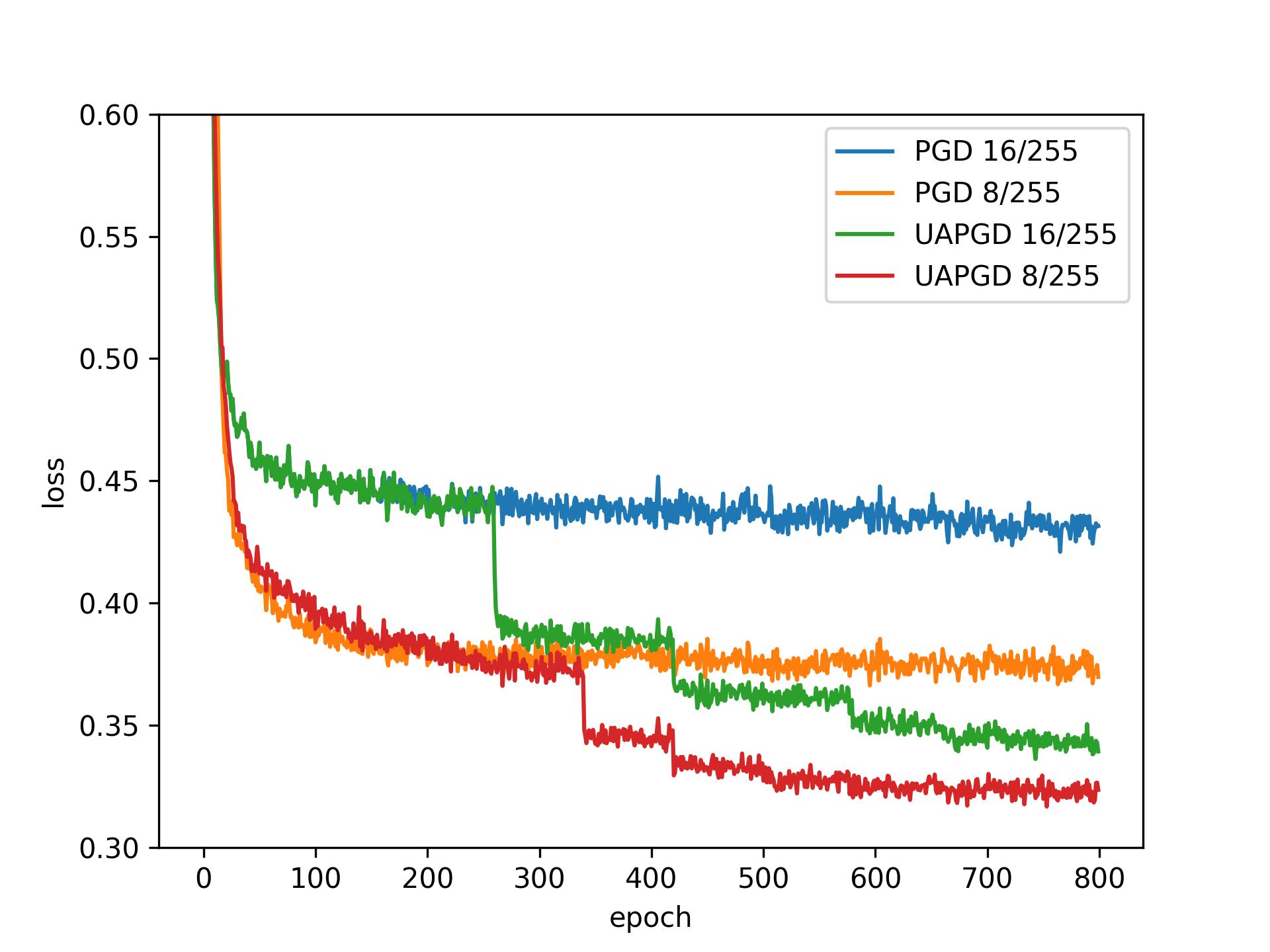}
       \caption{YOLOv3}
       \label{UAPGD_yolov3}
   \end{subfigure}
   \centering
   \begin{subfigure}{0.22\linewidth}
       \centering
       \includegraphics[width=1\linewidth]{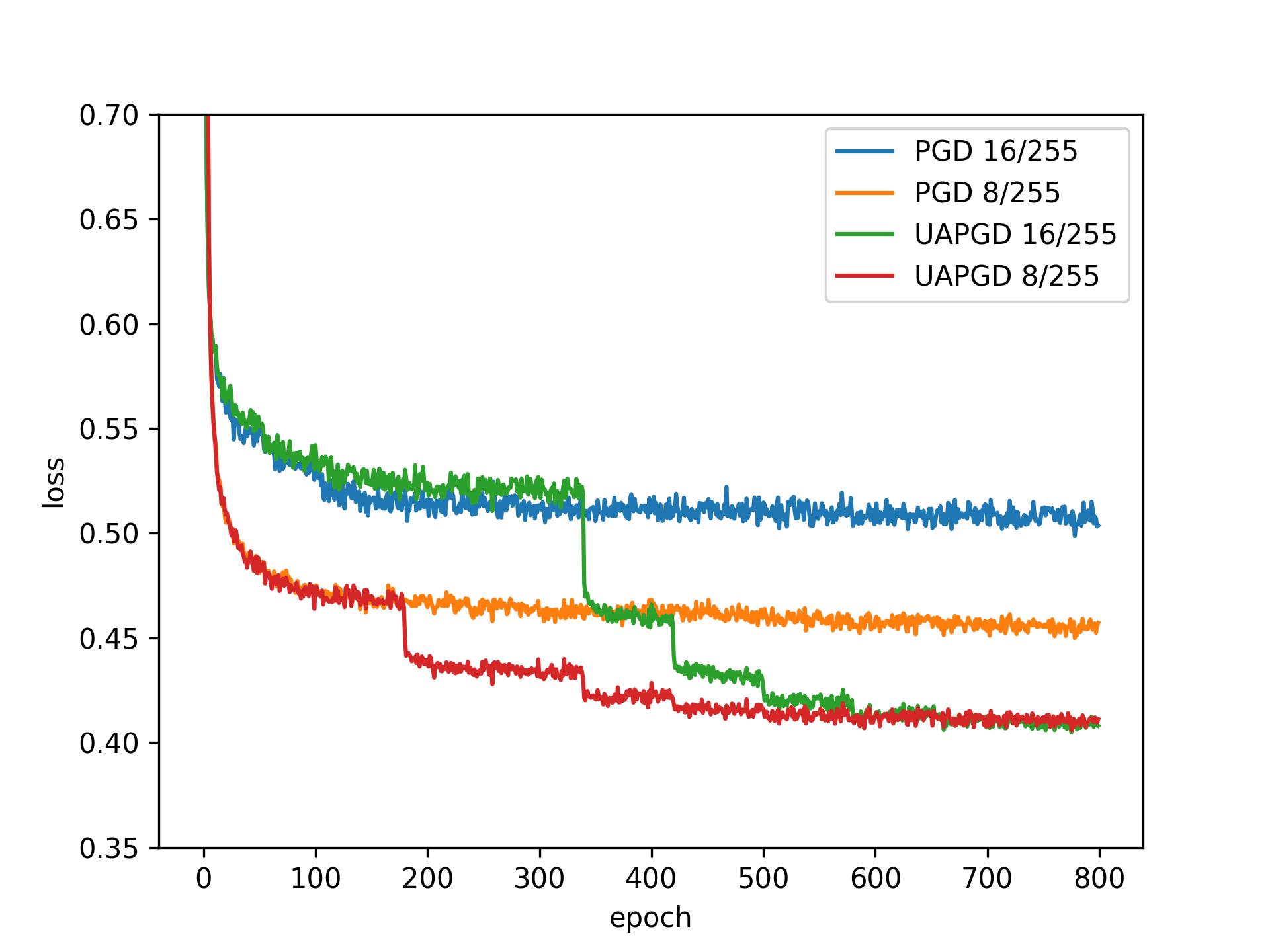}
       \caption{YOLOv5}
       \label{UAPGD_yolov5}
   \end{subfigure}
   \centering
   \begin{subfigure}{0.22\linewidth}
       \centering
       \includegraphics[width=1\linewidth]{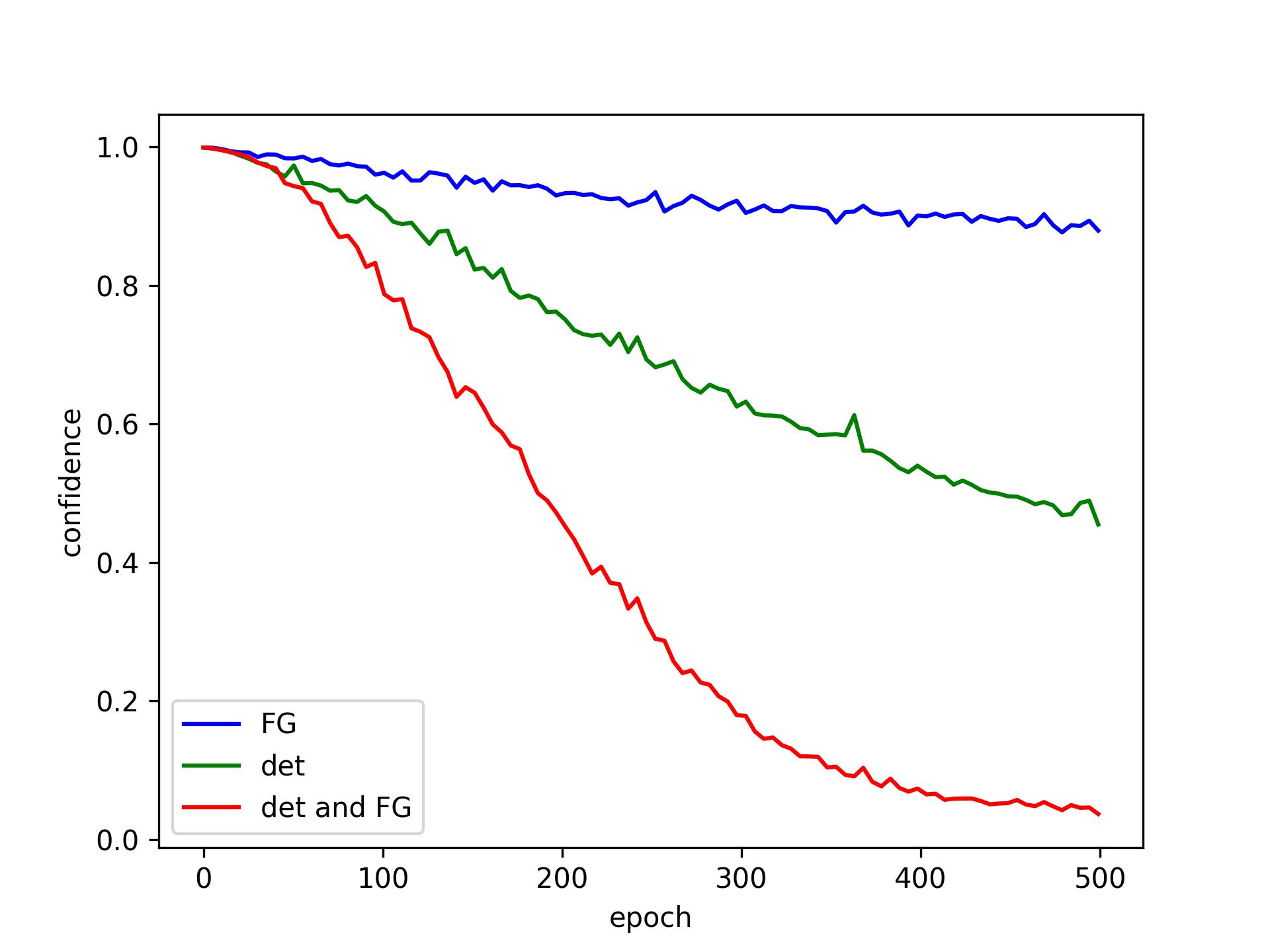}
       \caption{}
       \label{FG_exp1}
   \end{subfigure}
   \centering
   \begin{subfigure}{0.22\linewidth}
       \centering
       \includegraphics[width=1\linewidth]{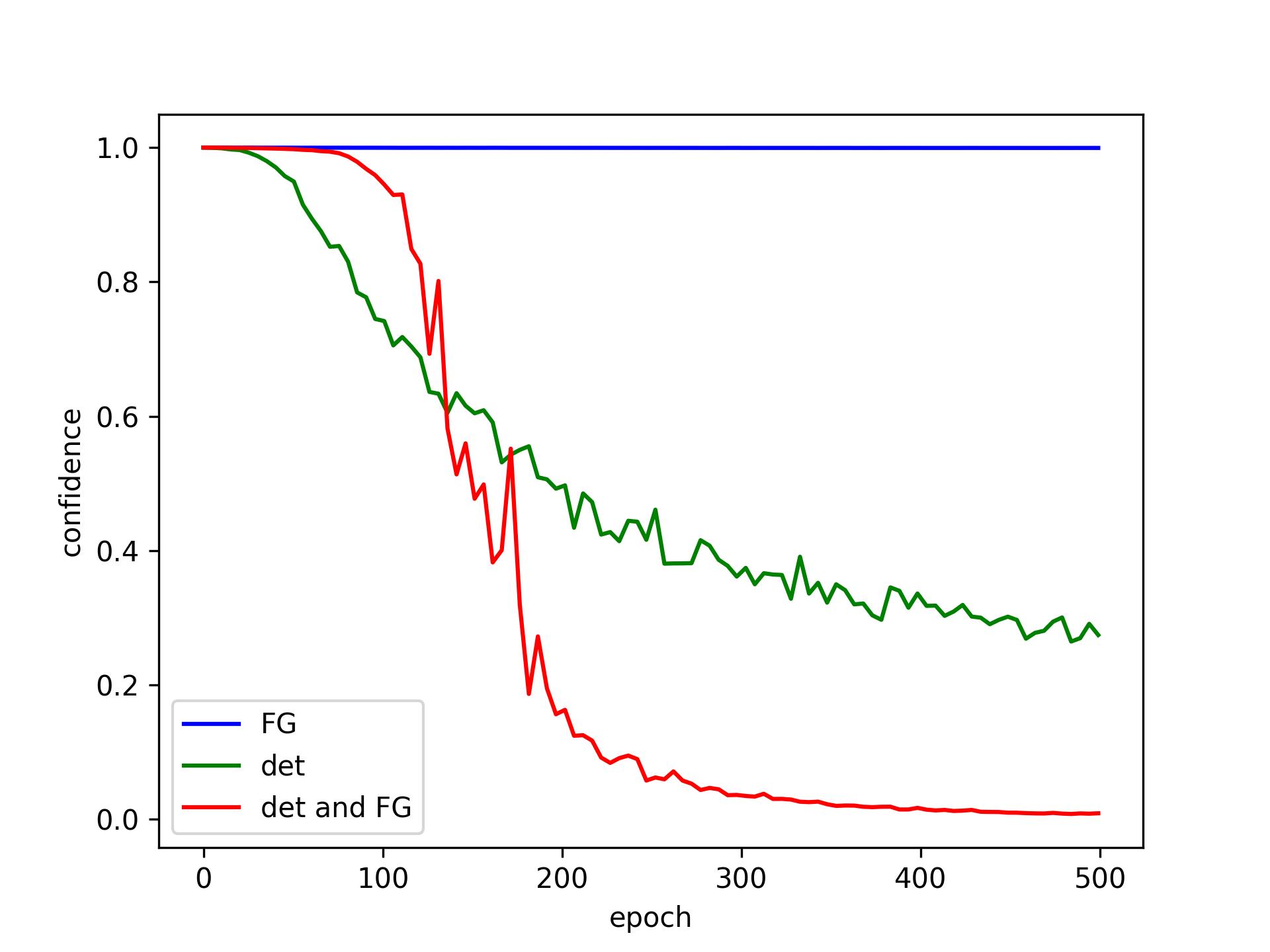}
       \caption{}
       \label{FG_exp2}
   \end{subfigure}
   \caption{(a) and (b) illustrate the variations of $L_{det}$, during the training process of generating adversarial triggers using PGD and UAPGD for attacking the YOLOv3 and YOLOv5 models. (c) and (d) present the confidence of some samples from the COCO dataset, outputted by the detector, during the training process of individual attacks. The blue, green, and red curved lines represent the utilization of $L_{FG}$, $L_{det}$, and both as the loss function respectively. For individual attacks, the optimization process is relatively easier so that step size decay is unnecessary.}
   \label{UAPGD_and_FG}
\end{figure}


As shown in \figureautorefname~\ref{digital result and trigger}, We compare our method with the more reasonable and advanced Dpatch2 ~\cite{lee2019physical}.
Considering the divergent optimization objectives, we substitute the maximization of the model loss function mentioned in ~\cite{lee2019physical} with the minimization of the proposed $L_{det}$ referred to as Dpatch2($L_{det}$). Undoubtedly, our method achieves the highest ASR in different experimental settings. 

A comparative analysis of the effectiveness between PGD and UAPGD 
reveals notable differences. Specifically, we observe that UAPGD, when employed with an identical attack step size as PGD, frequently results in a higher ASR.
Moreover, it can be observed that UAPGD yields better performance compared to adaptive step size optimizers like Adam.



\figureautorefname~\ref{UAPGD_yolov3} and \figureautorefname~\ref{UAPGD_yolov5} present a detailed depiction of the variations in the detection loss $L_{det}$ throughout the training process when generating adversarial triggers using both PGD and UAPGD techniques, specifically over YOLOv3 and YOLOv5 models. 
Notably, when operating under an equivalent number of epochs, UAPGD demonstrates a more pronounced capability to reduce $L_{det}$ compared to PGD. The effectiveness of the loss monitoring in UAPGD will be analyzed in detail in Appendix E.


Furthermore, as shown in \figureautorefname~\ref{digital result and trigger}, UAPGD exhibits a superior capability in generating smoother adversarial triggers. 
This increased smoothness can be attributed to the implementation of step size decay, which allows for a more refined optimization process in the generation of triggers. 
Consequently, adversarial triggers crafted via UAPGD display enhanced suitability for practical applications, such as printing and deployment in the physical world.



Based on results in \figureautorefname~\ref{digital result and trigger}
, we can observe that integrating $L_{FG}$ with suitable step size and weight into the loss function, coupled with using either PGD or UAPGD, can enhance the ASR. This improvement is particularly noticeable when applied to the YOLOv5 model. This observation highlights that employing appropriate feature guidance can improve the performance of universal attacks.
However, incorporating FG did not always achieve satisfactory results under certain weights and step sizes. 
This phenomenon is because in universal attacks, the inclusion of $L_{FG}$ introduces additional complexities to the optimization process. It requires appropriate adjustment of hyperparameters to fully leverage its effectiveness.
Hence, the optimal outcome can be achieved by ensemble UAPGD with it, which allows for finer control over the step size to facilitate the optimization process.
The empirical results indicate that the combination of UAPGD and FG in generating triggers yields the most effective attack performance.





To further investigate the potential role of Feature Guidance, we conducted individual adversarial attacks on each image with an adversarial trigger (fixed step size of 8/255, 500iteration) in the training dataset. We identified two primary roles of FG:
1, as shown in \figureautorefname~\ref{FG_exp1}, our observations reveal that FG can significantly influence the optimization direction during the training phase. 
    This influence is evidenced by the expedited convergence in adversarial trigger training, indicating the advantageous role of FG in enhancing training efficiency. 
%
2, As shown in \figureautorefname~\ref{FG_exp2}, when some images are challenging to attack with out-of-box adversarial triggers, introducing FG would enhance the attack effectiveness at this time.


\subsection{Experiment in simulator}

\begin{wraptable}{r}{0.4\textwidth}

    \begin{tabular}{cccccccccccccccccc}
    \specialrule{.1em}{.075em}{.075em} 
    attack method&Proportion\\ \hline
    No attack&0$\%$\\
    PGD&36.67$\%$\\
    UAPGD&45.56$\%$\\ 
    PGD,FG&46.67$\%$\\
    UAPGD,FG&58.89$\%$\\ 
    \specialrule{.1em}{.075em}{.075em} 
    \end{tabular}
    \caption{The proportion of time in the video duration during which the detector failed to recognize the stop sign.}
    \label{simasr}
\end{wraptable} 
In this section, we demonstrate the effects of our attack in autonomous driving scenarios, we use the Carla(0.9.14) simulator {~\cite{dosovitskiy2017carla}} to simulate the real driving process. 
In the simulator, we utilize the YOLOv3 model as the detection network.


Due to the significant differences between images from the COCO dataset and those generated by the Carla simulator, the adversarial trigger developed in Section \ref{mainre} is not directly applicable to Carla. 
To tailor an adversarial trigger specifically for the Carla simulator, we compile a training dataset comprising 242 images featuring stop signs from the simulator. 
The process of attacking the Carla simulator is conducted in four distinct manners: employing either PGD or UAPGD, combined with a loss function of either 
$L_{det}+\lambda_{tv}L_{tv}$ or $L_{all}$, where $\lambda_{FG}=0.1$ and the initial step size is 16/255.

\begin{wrapfigure}{r}{0.4\textwidth}
    \centering
    \includegraphics[width=0.9\linewidth]{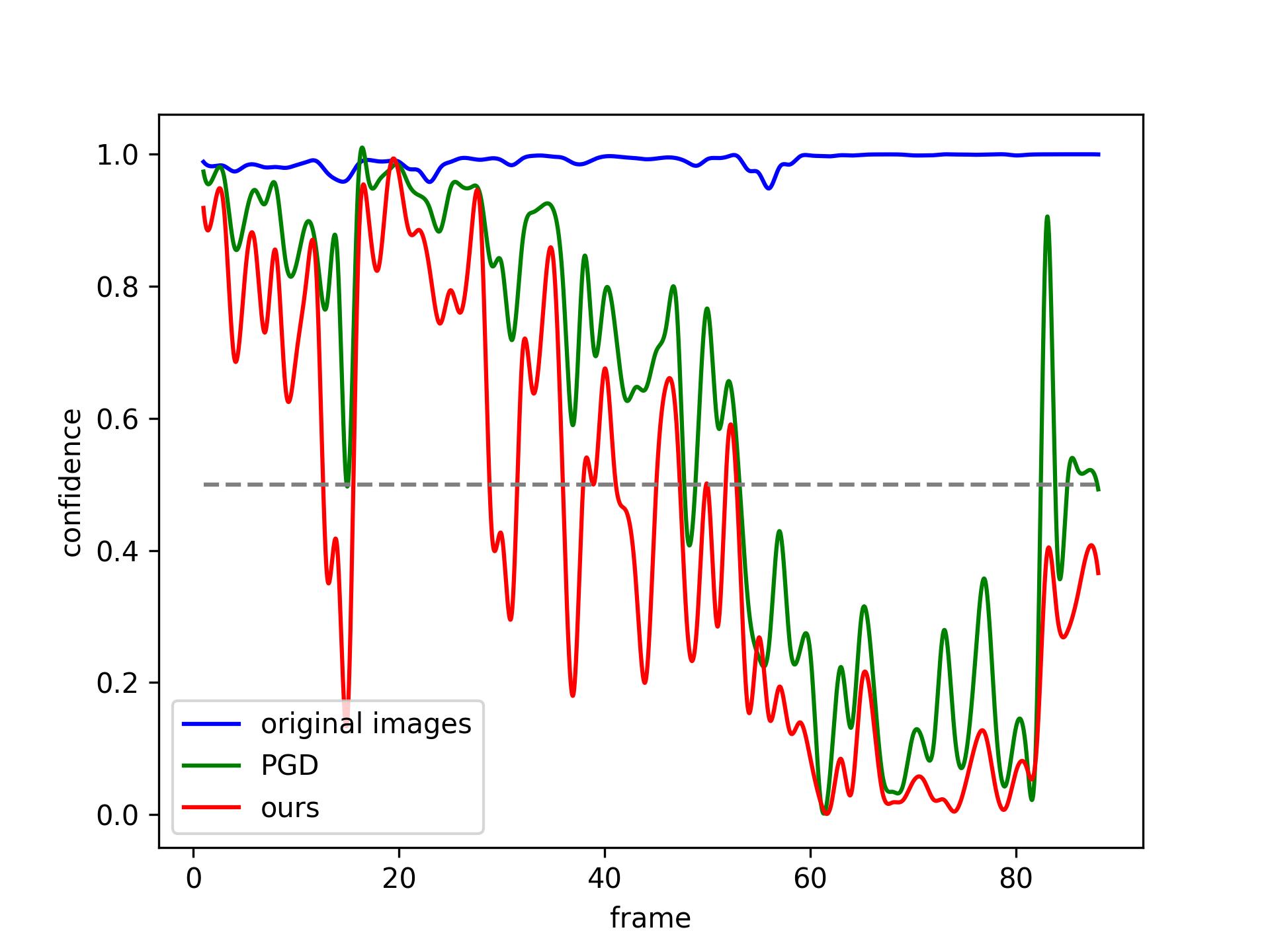}
    \captionof{figure}{ We uniformly extracted approximately 90 frames from the 9s video, which is recorded by the simulated vehicle's onboard camera, to showcase the confidence variations of the detector's predictions} 
    \label{conf}
\end{wrapfigure}

To evaluate the effectiveness of the attack, we implement a procedure where the adversarial trigger is placed under the stop sign in the simulator. 
Subsequently, a vehicle is navigated towards this stop sign, and a 9-second video is recorded using an onboard camera. 
The key metric is the percentage of time during which the detection system fails to recognize the stop sign. 


    



As shown in \tableautorefname~\ref{simasr}, without any adversarial trigger, the car consistently recognizes the stop sign. 
When the attack is initiated using the PGD method alone, 
the attack duration is approximately 3 to 4 seconds. 
However, the integration of either UAPGD or FG methods into the attack strategy results in a slight increase in the attack duration by roughly an additional second. 
Notably, the implementation of an ensemble approach, which combines these methods, further enhances the attack performance, prolonging the duration to over 5 seconds. 



Moreover, to provide a more direct illustration of the attack impact, we have plotted the trend of the stop sign detection confidence as the car approaches it, shown as in \figureautorefname~\ref{conf}. 


The empirical results reveal that the efficacy of our attack method consistently surpasses that of the vanilla PGD approach. 
Particularly in the final 5 seconds of the video, during which the adversarial trigger persistently causes the detection system to lose sight of the stop sign. 
This effect illustrates the potential of our method to compromise the functionality of autonomous driving systems.

\begin{figure}[!t]
    \captionsetup[subfigure]{labelformat=empty}
    \centering
    \begin{subfigure}{0.16\linewidth}
        \centering
        \includegraphics[width=0.9\linewidth]{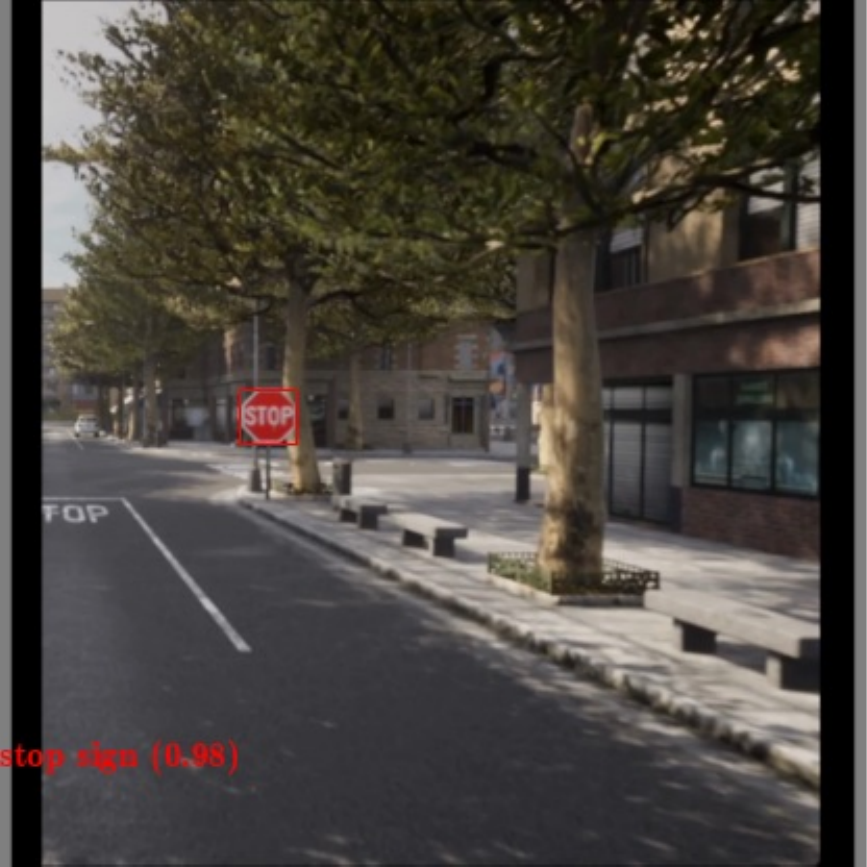}
        \caption{$conf=0.98$}
        \label{origin1}
    \end{subfigure}
    \centering
    \begin{subfigure}{0.16\linewidth}
        \centering
        \includegraphics[width=0.9\linewidth]{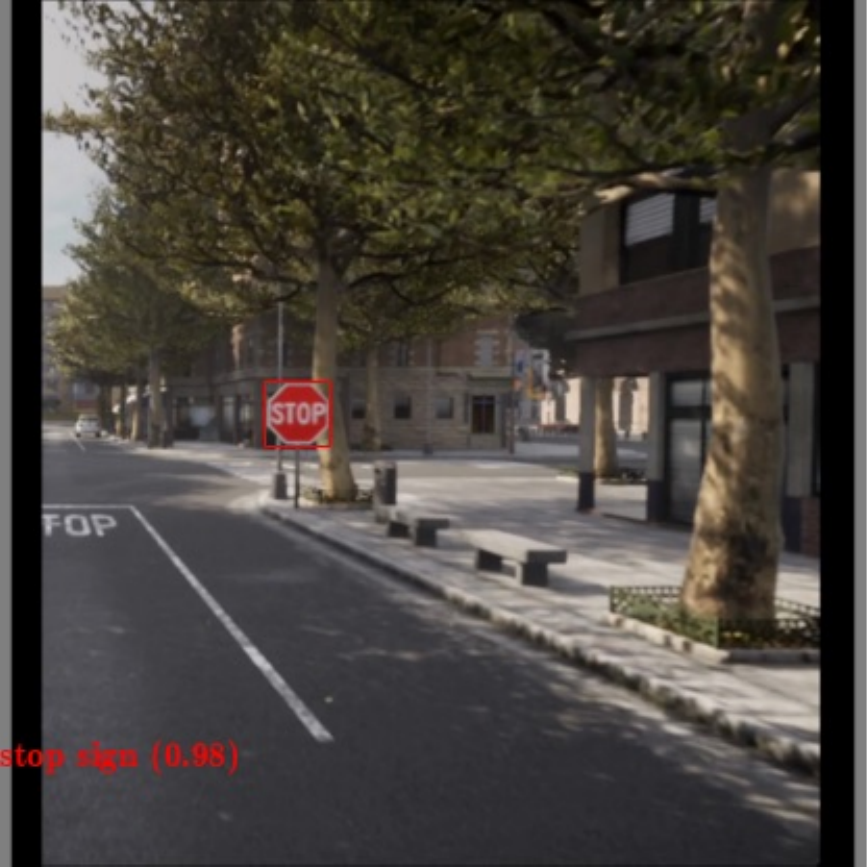}
        \caption{$conf=0.98$}
        \label{origin1}
    \end{subfigure}
    \centering
    \begin{subfigure}{0.16\linewidth}
        \centering
        \includegraphics[width=0.9\linewidth]{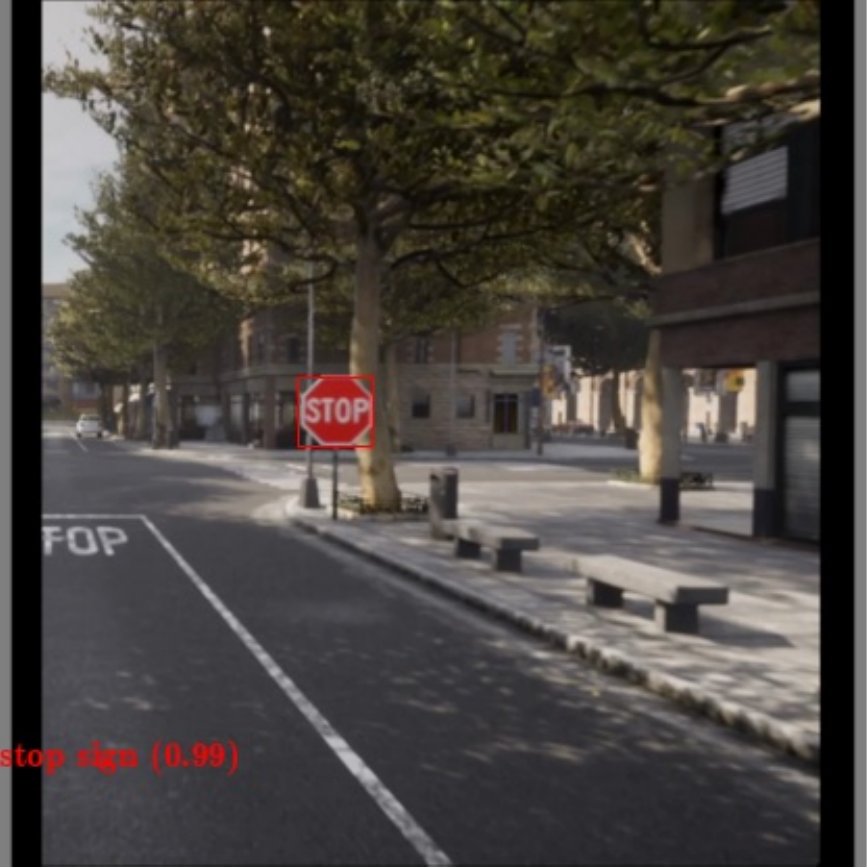}
        \caption{$conf=0.99$}
        \label{origin1}
    \end{subfigure}
    \centering
    \begin{subfigure}{0.16\linewidth}
        \centering
        \includegraphics[width=0.9\linewidth]{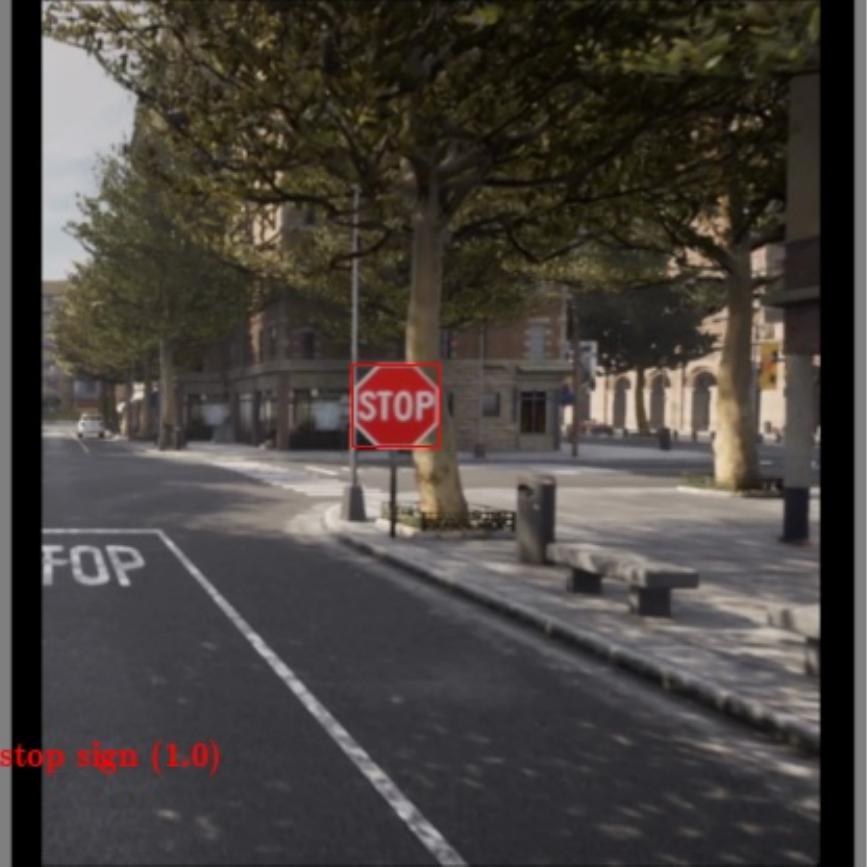}
        \caption{$conf=1.0$}
        \label{origin1}
    \end{subfigure}
    \centering
    \begin{subfigure}{0.16\linewidth}
        \centering
        \includegraphics[width=0.9\linewidth]{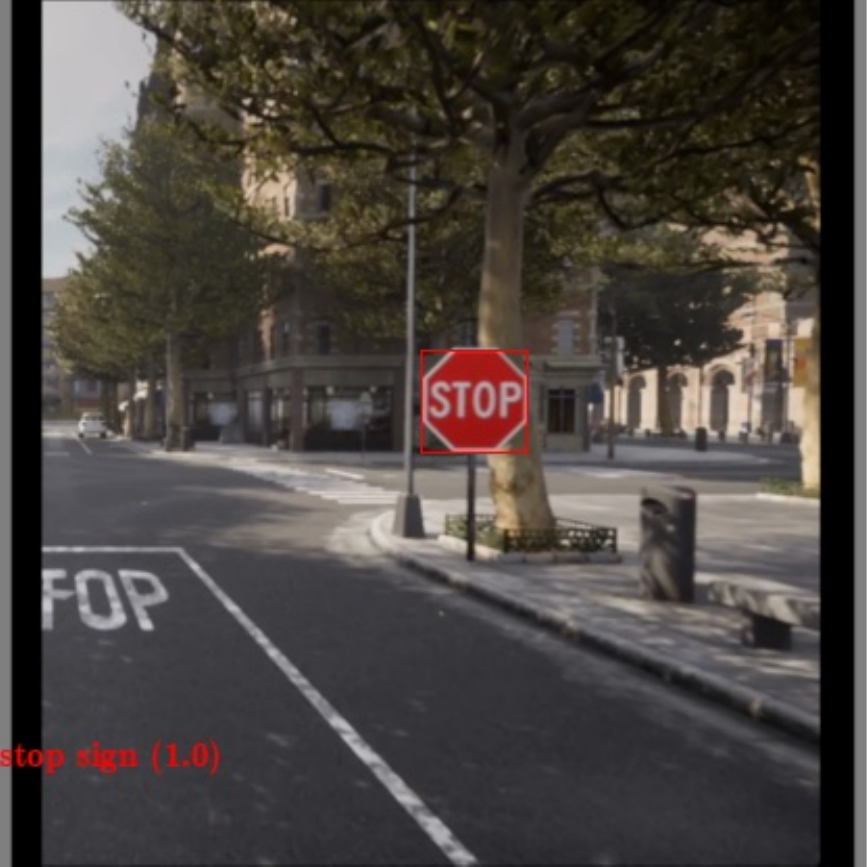}
        \caption{$conf=1.0$}
        \label{origin1}
    \end{subfigure}
    \centering
    \begin{subfigure}{0.16\linewidth}
        \centering
        \includegraphics[width=0.9\linewidth]{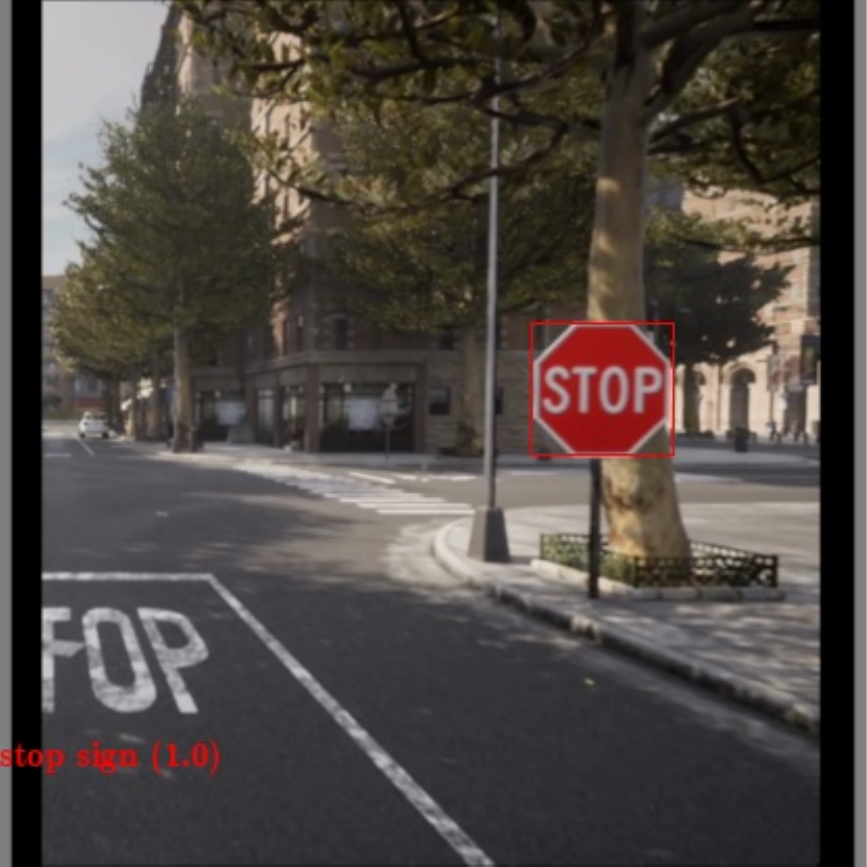}
        \caption{$conf=1.0$}
        \label{origin1}
    \end{subfigure}
    
    \centering
    \begin{subfigure}{0.16\linewidth}
        \centering
        \includegraphics[width=0.9\linewidth]{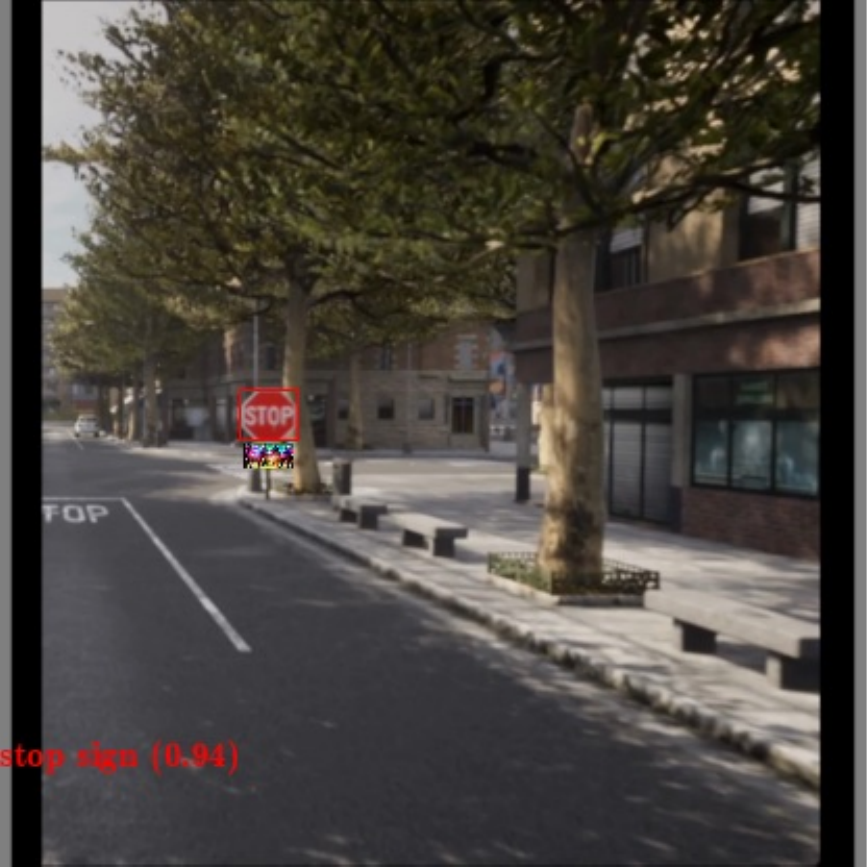}
        \caption{$conf=0.94$}
        \label{origin1}
    \end{subfigure}
    \centering
    \begin{subfigure}{0.16\linewidth}
        \centering
        \includegraphics[width=0.9\linewidth]{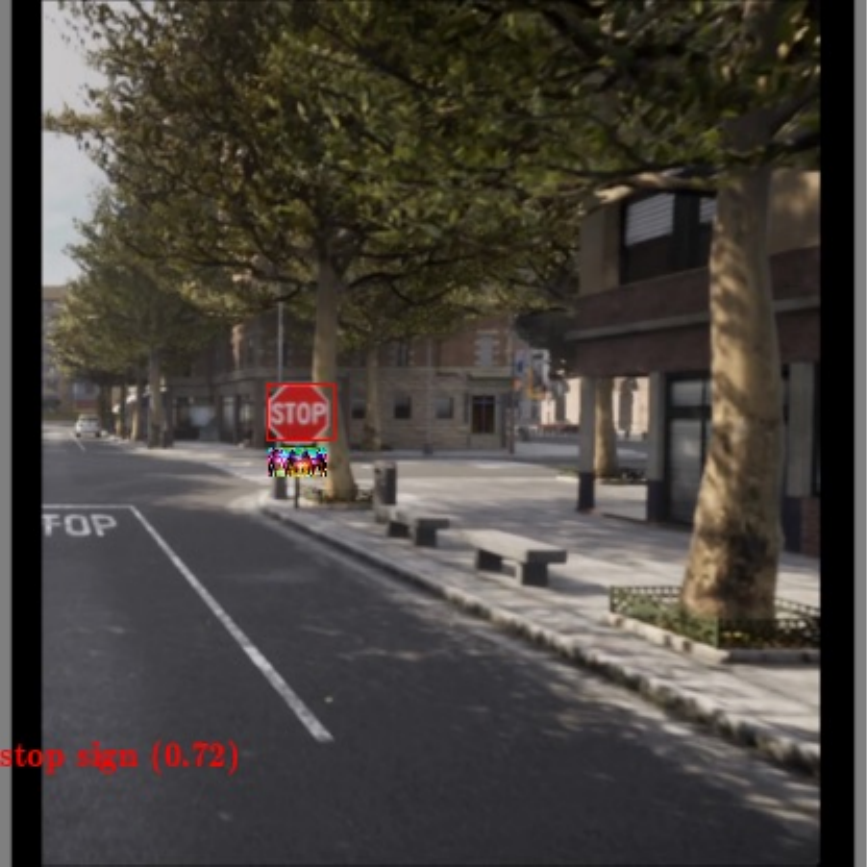}
        \caption{$conf=0.72$}
        \label{origin1}
    \end{subfigure}
    \centering
    \begin{subfigure}{0.16\linewidth}
        \centering
        \includegraphics[width=0.9\linewidth]{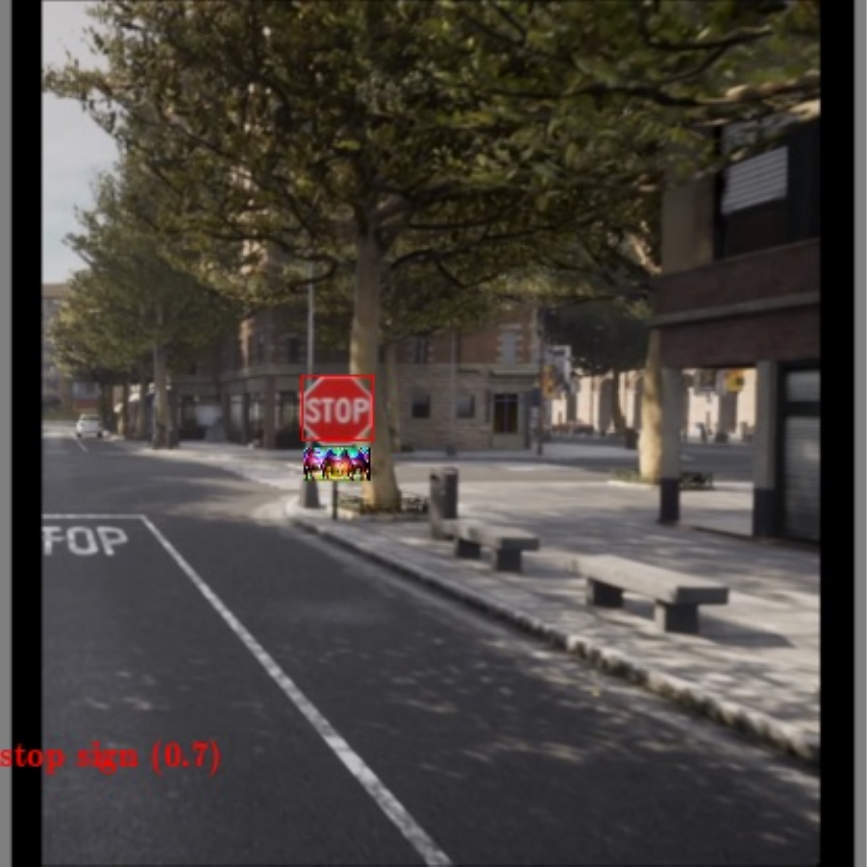}
        \caption{$conf=0.70$}
        \label{origin1}
    \end{subfigure}
    \centering
    \begin{subfigure}{0.16\linewidth}
        \centering
        \includegraphics[width=0.9\linewidth]{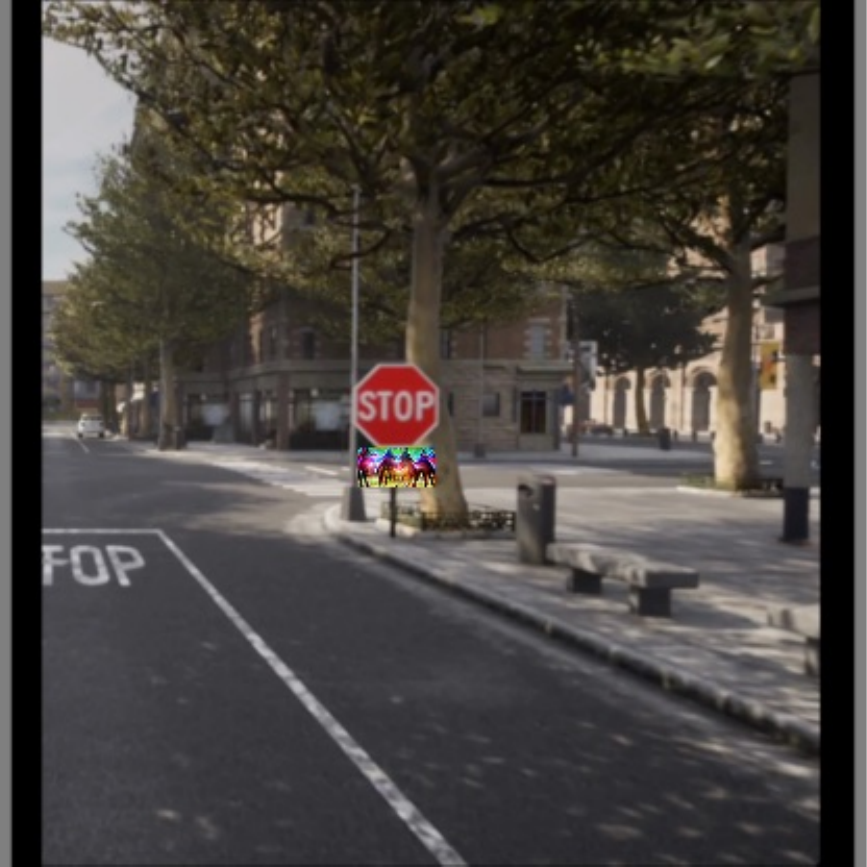}
        \caption{{\color{red}$conf<0.5$}}
        \label{origin1}
    \end{subfigure}
    \centering
    \begin{subfigure}{0.16\linewidth}
        \centering
        \includegraphics[width=0.9\linewidth]{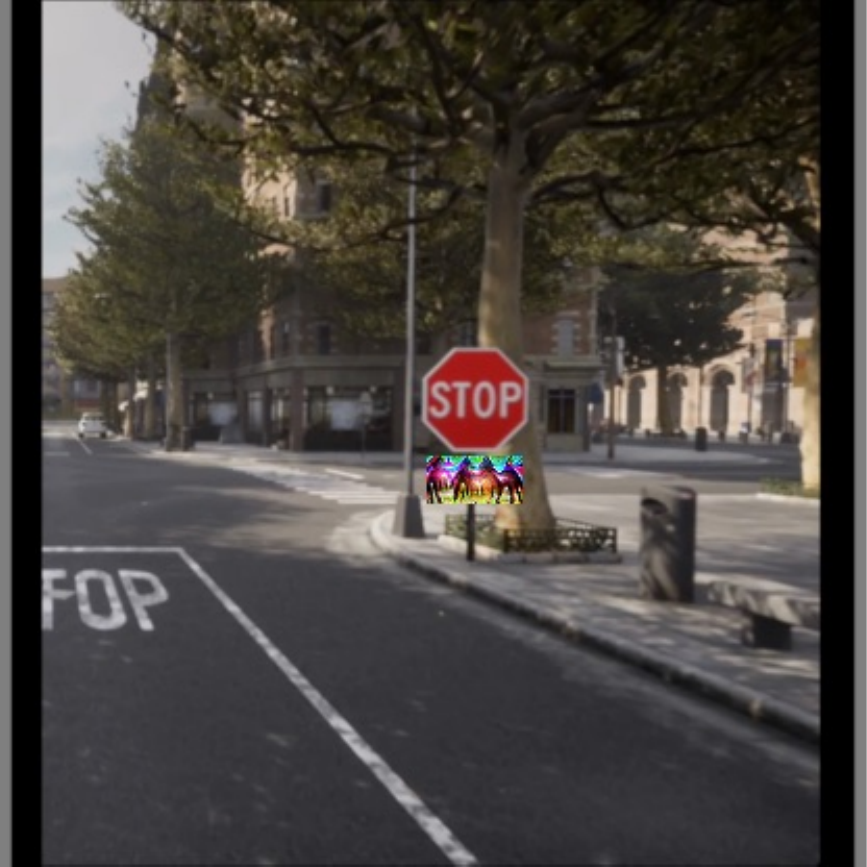}
        \caption{{\color{red}$conf<0.5$}}
        \label{origin1}
    \end{subfigure}
    \centering
    \begin{subfigure}{0.16\linewidth}
        \centering
        \includegraphics[width=0.9\linewidth]{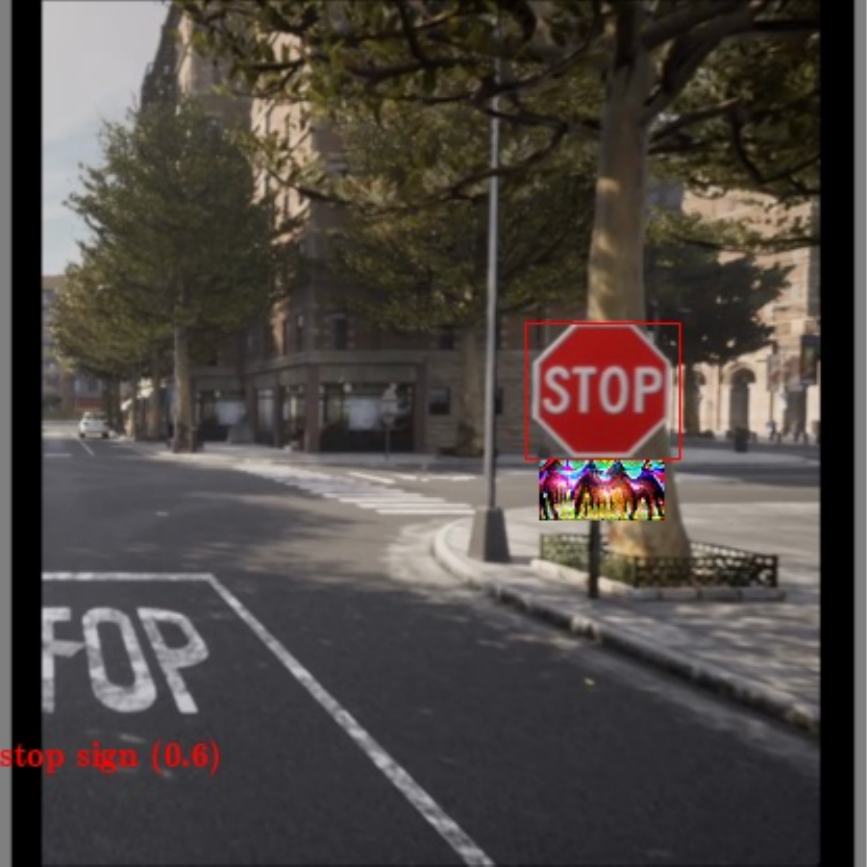}
        \caption{$conf=0.6$}
        \label{origin1}
    \end{subfigure}

    \centering
    \begin{subfigure}{0.16\linewidth}
        \centering
        \includegraphics[width=0.9\linewidth]{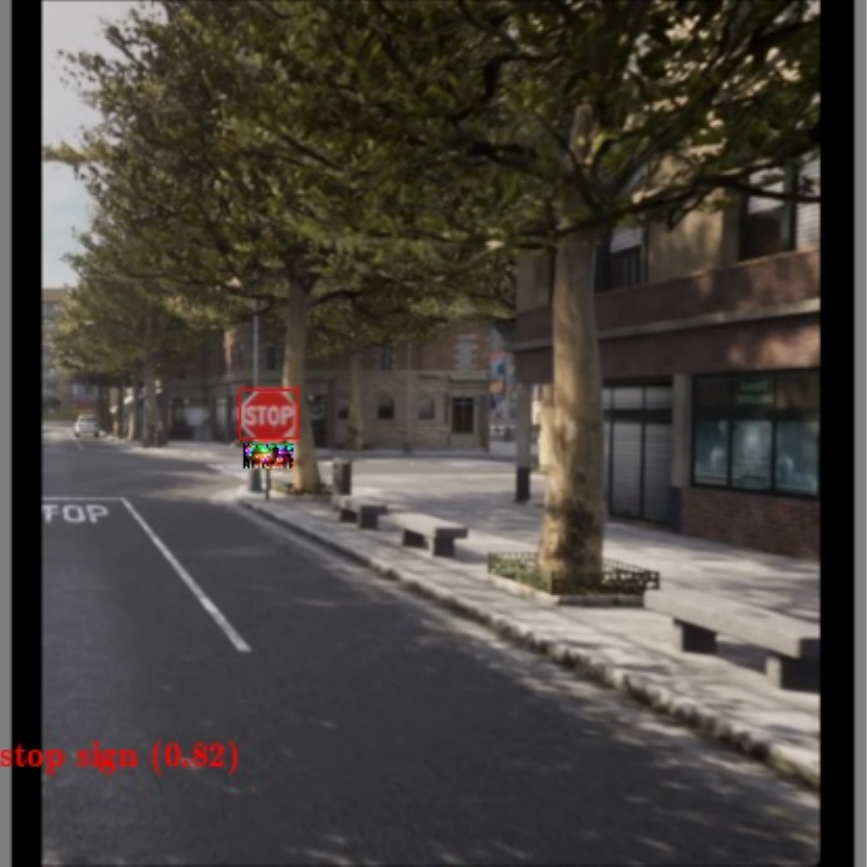}
        \caption{{\color{red}$conf=0.82$}}
        \label{origin1}
    \end{subfigure}
    \centering
    \begin{subfigure}{0.16\linewidth}
        \centering
        \includegraphics[width=0.9\linewidth]{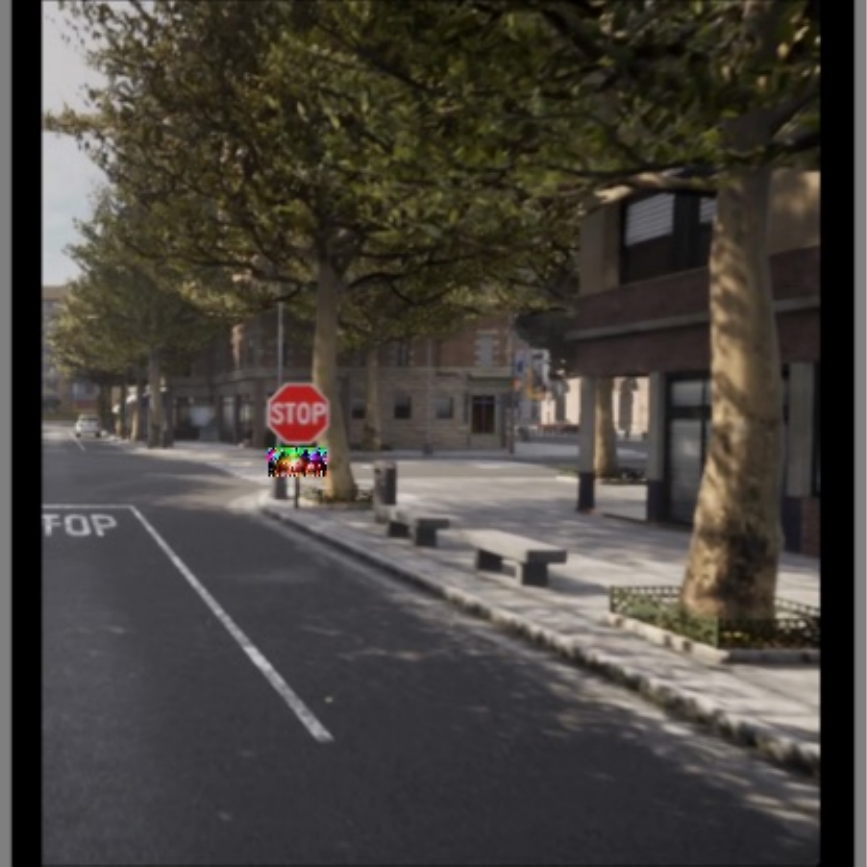}
        \caption{{\color{red}$conf<0.5$}}
        \label{origin1}
    \end{subfigure}
    \centering
    \begin{subfigure}{0.16\linewidth}
        \centering
        \includegraphics[width=0.9\linewidth]{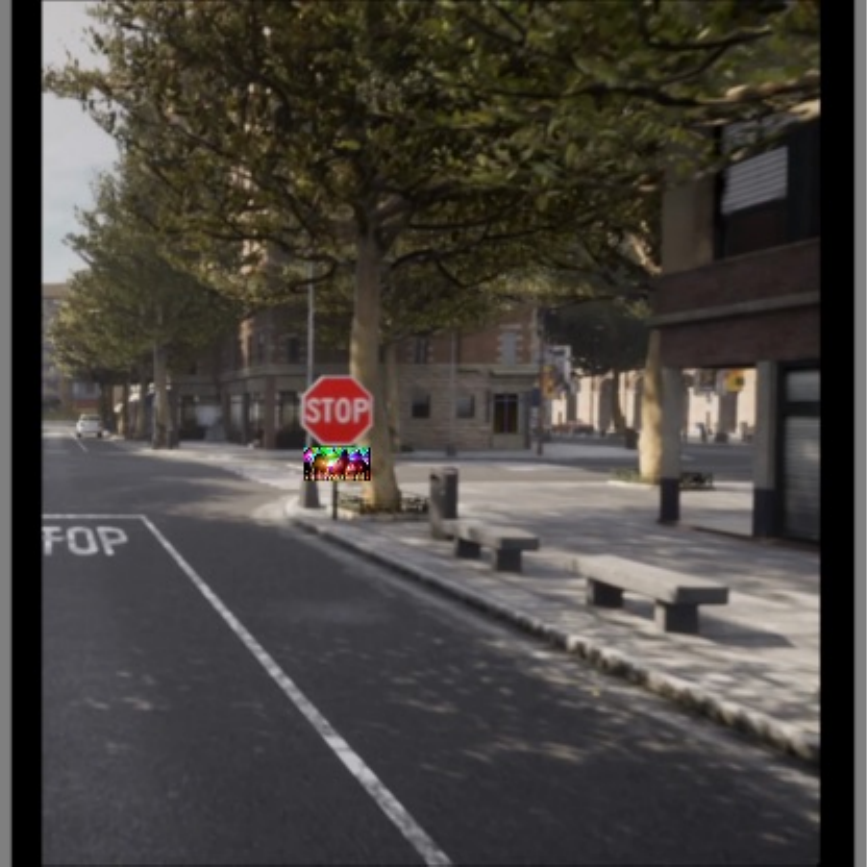}
        \caption{{\color{red}$conf<0.5$}}
        \label{origin1}
    \end{subfigure}
    \centering
    \begin{subfigure}{0.16\linewidth}
        \centering
        \includegraphics[width=0.9\linewidth]{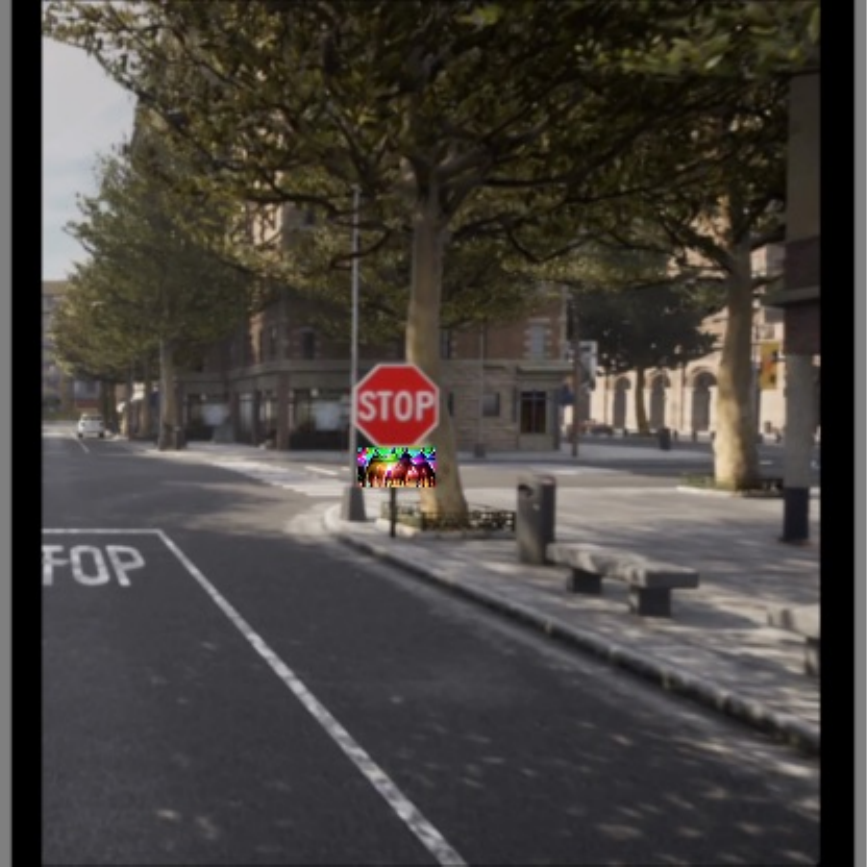}
        \caption{{\color{red}$conf<0.5$}}
        \label{origin1}
    \end{subfigure}
    \centering
    \begin{subfigure}{0.16\linewidth}
        \centering
        \includegraphics[width=0.9\linewidth]{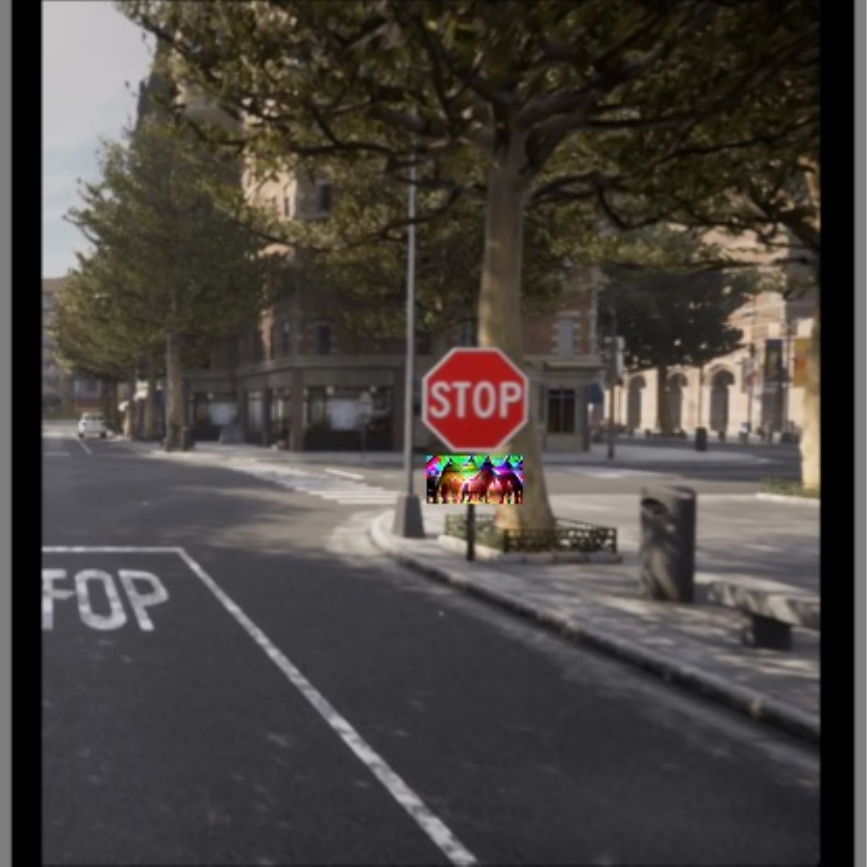}
        \caption{{\color{red}$conf<0.5$}}
        \label{origin1}
    \end{subfigure}
    \centering
    \begin{subfigure}{0.16\linewidth}
        \centering
        \includegraphics[width=0.9\linewidth]{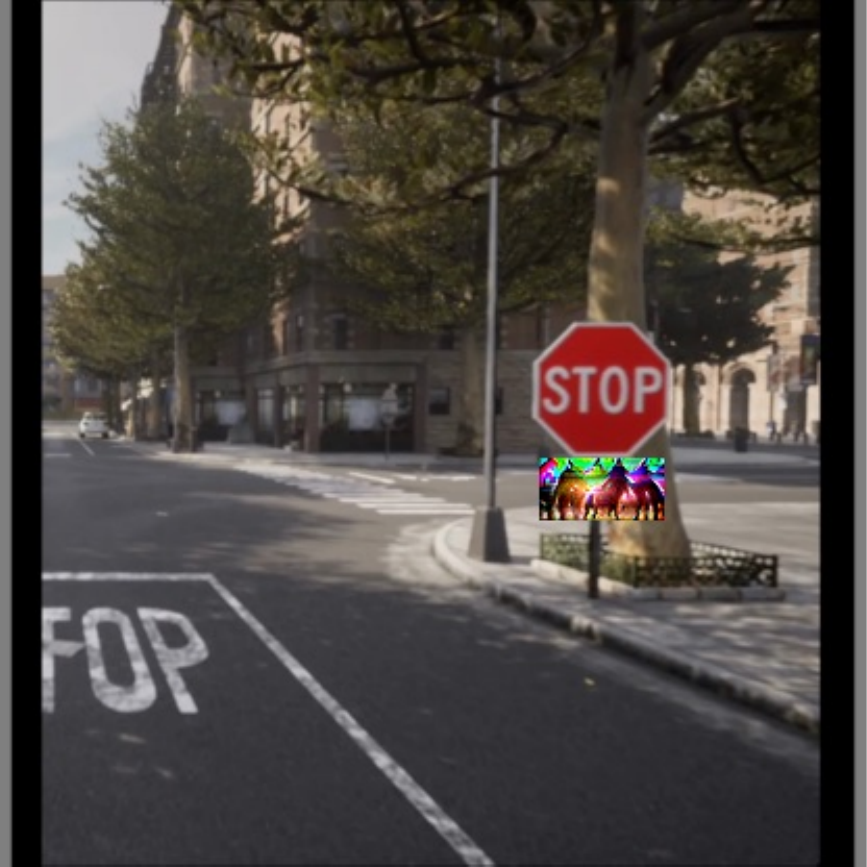}
        \caption{{\color{red}$conf<0.5$}}
        \label{origin1}
    \end{subfigure}
    \caption{ Six frames are extracted uniformly from the video. Row 1 represents a clean stop sign with the detector's output. In Row 2 and Row 3, we deploy the trigger generated by the PGD and ensemble method respectively.
    }
    \label{simulator}
\end{figure}

As shown in the \figureautorefname~\ref{simulator}, we extract six frames uniformly from the video as examples to demonstrate the effects of this attack in driving scenarios.

The observations imply that normally the car will always recognize the stop sign (Row 1 of \figureautorefname~\ref{simulator}, each stop is marked by a red box). However, after the adversarial trigger created by our attack method (Row 3) is added to the stop sign, the car fails to recognize the stop sign(no red box). Compared with the attack effect of the adversarial trigger generated by PGD (Row 2), it can be seen that our method is clearly superior to the PGD. 
It is worth mentioning that the perturbation generated by Dpatch2 is suboptimal in the simulator, as demonstrated in Appendix F.

\subsection{Robustness of our attack in real world}
\begin{figure}[!tb]
   \centering
   \begin{subfigure}{0.16\linewidth}
       \centering
       \includegraphics[width=0.83\linewidth]{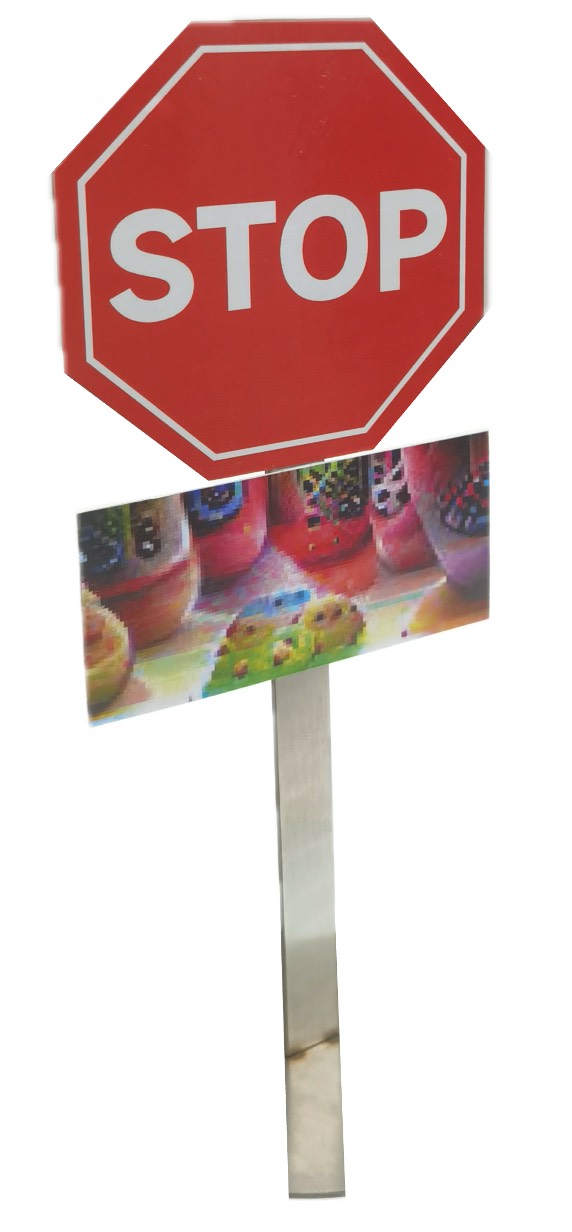}
       \caption*{}
       \label{show}
   \end{subfigure}
   \centering
   \begin{subfigure}{0.4\linewidth}
       \centering
       \includegraphics[width=1\linewidth]{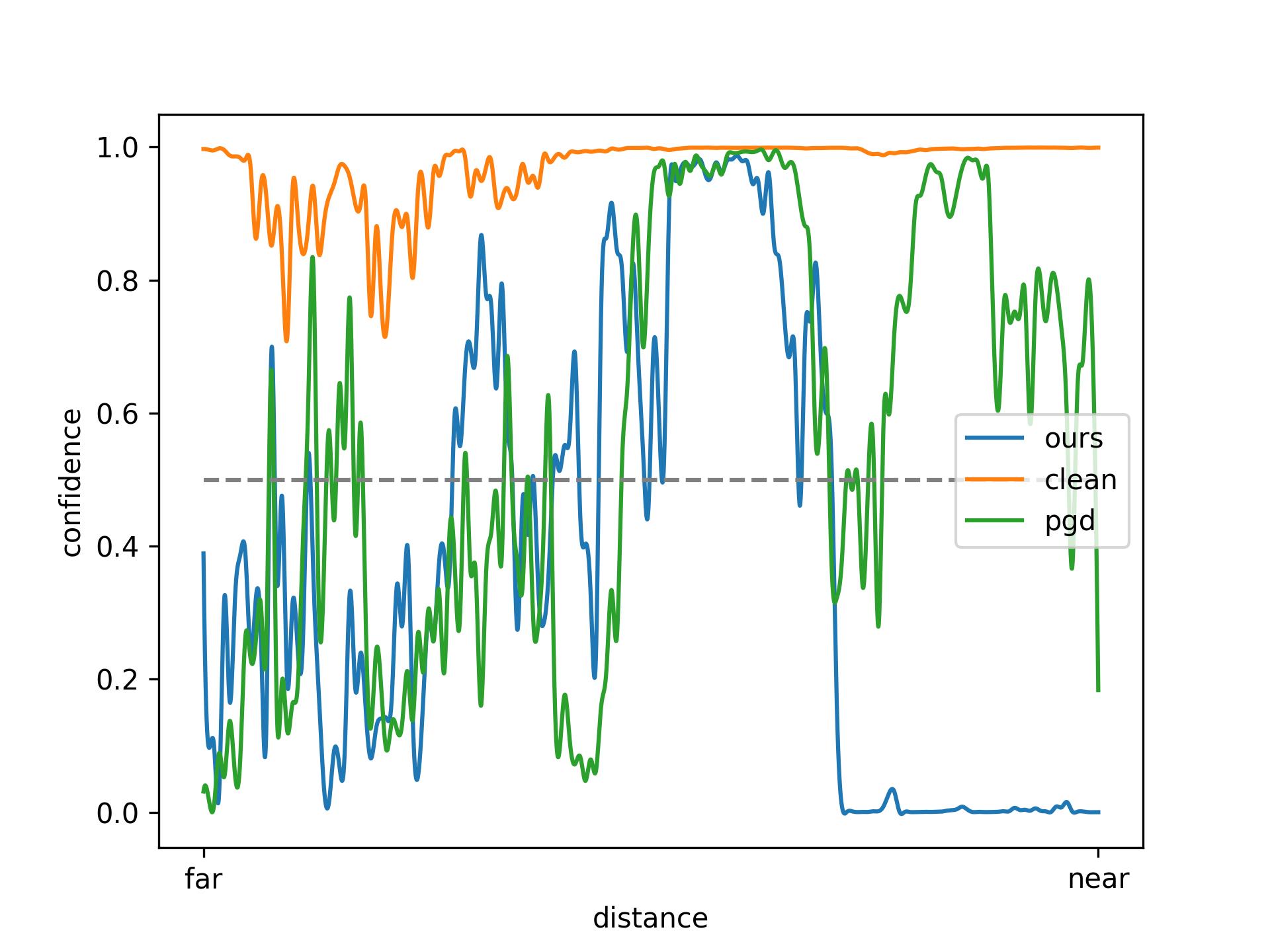}
       \caption{}
       \label{far}
   \end{subfigure}
   \centering
   \begin{subfigure}{0.4\linewidth}
       \centering
       \includegraphics[width=1\linewidth]{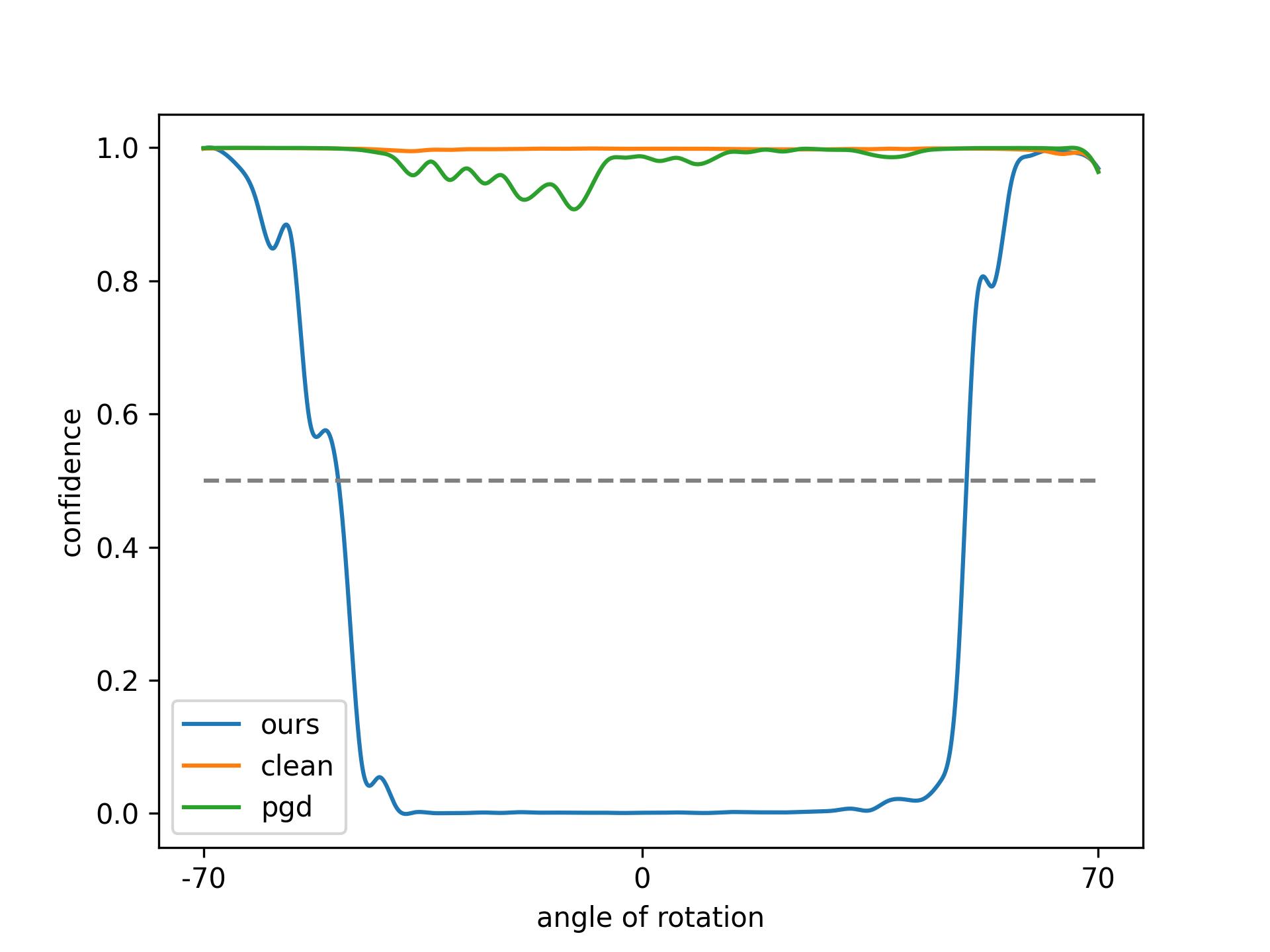}
       \caption{}
       \label{rotate}
   \end{subfigure}
   
   \caption{ We print and deploy the adversarial trigger in the physical world. (a) illustrates the confidence variation of the stop sign as the distance increases. (b) depicts the confidence variation of the stop sign when facing different angles changing.}
   \label{realworld}
   
\end{figure}
To observe the attack effectiveness and robustness of the universal adversarial trigger obtained through  \algorithmautorefname~\ref{algo_disjdecomp} in the real world, we print out the adversarial trigger and post it under a stop sign. Moreover, in the case where clean images serve as a reference, we compare our method with the normal PGD.

We examined the variation trends in the confidence of the detector's output for recognizing the stop sign using an onboard camera under different conditions: as the distance between the onboard camera and the stop sign decreased from far to near, and as the onboard camera's perspective towards the stop sign shifted from -70-degree side view to a 70-degree side view. The real-world experiment results are shown in \figureautorefname~\ref{realworld}.




According to \figureautorefname~\ref{far}, due to the relatively long distance between the vehicle and the stop position in the first half of the curve, the trigger occupies a small proportion of the field of view, which diminishes the stability of successful attacks. In the second half of the curve, where the distance between the stop and the vehicle is closer, the attacks become more stable. Our method consistently maintains a higher success rate in this scenario. Examining \figureautorefname~\ref{rotate}, it can be observed that when facing the stop sign directly, our method misleads the detector to output a remarkably low confidence. As the angle approaches approximately $\pm50$ degrees, our method becomes ineffective. In contrast, the triggers generated by PGD are not smooth, which prevents them from effectively imprinting adversarial features into the real world. 

Overall, in the real world, our attack consistently exhibits robust and effective performance when the onboard camera is approaching a stop sign. This means that our attack can successfully mislead an autonomous driving system to disregard the stop sign unconsciously.
\section{Conclusion}

In this work, we present a novel adversarial attack methodology that extends beyond the bounding boxes of objects. 
This method poses greater challenges compared to previous studies due to its tendency to focus on less informative areas. 
Thus, we propose the FG techniques and UAPGD optimization strategy, specifically designed to augment the efficacy of this attack in both digital and physical realms. 
Furthermore, we validate the effectiveness of our attack method through its application in autonomous driving scenarios and real-world conditions. 
Discussion of limitations and potential negative impact of our works will be placed in Appendix G and immediately following that we also briefly address future work.

\bibliographystyle{splncs04}
\bibliography{main}

\begin{thebibliography}{10}
\providecommand{\url}[1]{\texttt{#1}}
\providecommand{\urlprefix}{URL }
\providecommand{\doi}[1]{https://doi.org/#1}

\bibitem{aich2023leveraging}
Aich, A., Li, S., Song, C., Asif, M.S., Krishnamurthy, S.V., Roy-Chowdhury, A.K.: Leveraging local patch differences in multi-object scenes for generative adversarial attacks. In: Proceedings of the IEEE/CVF Winter Conference on Applications of Computer Vision. pp. 1308--1318 (2023)

\bibitem{athalye2018synthesizing}
Athalye, A., Engstrom, L., Ilyas, A., Kwok, K.: Synthesizing robust adversarial examples. In: International conference on machine learning. pp. 284--293. PMLR (2018)

\bibitem{brendel2017decision}
Brendel, W., Rauber, J., Bethge, M.: Decision-based adversarial attacks: Reliable attacks against black-box machine learning models. arXiv preprint arXiv:1712.04248  (2017)

\bibitem{cao2019adversarial}
Cao, Y., Xiao, C., Cyr, B., Zhou, Y., Park, W., Rampazzi, S., Chen, Q.A., Fu, K., Mao, Z.M.: Adversarial sensor attack on lidar-based perception in autonomous driving. In: Proceedings of the 2019 ACM SIGSAC conference on computer and communications security. pp. 2267--2281 (2019)

\bibitem{carlini2017towards}
Carlini, N., Wagner, D.: Towards evaluating the robustness of neural networks. In: 2017 ieee symposium on security and privacy (sp). pp. 39--57. Ieee (2017)

\bibitem{chen2015deepdriving}
Chen, C., Seff, A., Kornhauser, A., Xiao, J.: Deepdriving: Learning affordance for direct perception in autonomous driving. In: Proceedings of the IEEE international conference on computer vision. pp. 2722--2730 (2015)

\bibitem{chen2020hopskipjumpattack}
Chen, J., Jordan, M.I., Wainwright, M.J.: Hopskipjumpattack: A query-efficient decision-based attack. In: 2020 ieee symposium on security and privacy (sp). pp. 1277--1294. IEEE (2020)

\bibitem{chen2017zoo}
Chen, P.Y., Zhang, H., Sharma, Y., Yi, J., Hsieh, C.J.: Zoo: Zeroth order optimization based black-box attacks to deep neural networks without training substitute models. In: Proceedings of the 10th ACM workshop on artificial intelligence and security. pp. 15--26 (2017)

\bibitem{croce2020reliable}
Croce, F., Hein, M.: Reliable evaluation of adversarial robustness with an ensemble of diverse parameter-free attacks. In: International conference on machine learning. pp. 2206--2216. PMLR (2020)

\bibitem{deng2019arcface}
Deng, J., Guo, J., Xue, N., Zafeiriou, S.: Arcface: Additive angular margin loss for deep face recognition. In: Proceedings of the IEEE/CVF conference on computer vision and pattern recognition. pp. 4690--4699 (2019)

\bibitem{deng2019retinaface}
Deng, J., Guo, J., Zhou, Y., Yu, J., Kotsia, I., Zafeiriou, S.: Retinaface: Single-stage dense face localisation in the wild. arXiv preprint arXiv:1905.00641  (2019)

\bibitem{dong2018boosting}
Dong, Y., Liao, F., Pang, T., Su, H., Zhu, J., Hu, X., Li, J.: Boosting adversarial attacks with momentum. In: Proceedings of the IEEE conference on computer vision and pattern recognition. pp. 9185--9193 (2018)

\bibitem{dong2019evading}
Dong, Y., Pang, T., Su, H., Zhu, J.: Evading defenses to transferable adversarial examples by translation-invariant attacks. In: Proceedings of the IEEE/CVF Conference on Computer Vision and Pattern Recognition. pp. 4312--4321 (2019)

\bibitem{dosovitskiy2017carla}
Dosovitskiy, A., Ros, G., Codevilla, F., Lopez, A., Koltun, V.: Carla: An open urban driving simulator. In: Conference on robot learning. pp. 1--16. PMLR (2017)

\bibitem{duan2020adversarial}
Duan, R., Ma, X., Wang, Y., Bailey, J., Qin, A.K., Yang, Y.: Adversarial camouflage: Hiding physical-world attacks with natural styles. In: Proceedings of the IEEE/CVF conference on computer vision and pattern recognition. pp. 1000--1008 (2020)

\bibitem{eykholt2018robust}
Eykholt, K., Evtimov, I., Fernandes, E., Li, B., Rahmati, A., Xiao, C., Prakash, A., Kohno, T., Song, D.: Robust physical-world attacks on deep learning visual classification. In: Proceedings of the IEEE conference on computer vision and pattern recognition. pp. 1625--1634 (2018)

\bibitem{goodfellow2014explaining}
Goodfellow, I.J., Shlens, J., Szegedy, C.: Explaining and harnessing adversarial examples. arXiv preprint arXiv:1412.6572  (2014)

\bibitem{grigorescu2020survey}
Grigorescu, S., Trasnea, B., Cocias, T., Macesanu, G.: A survey of deep learning techniques for autonomous driving. Journal of Field Robotics  \textbf{37}(3),  362--386 (2020)

\bibitem{hoory2020dynamic}
Hoory, S., Shapira, T., Shabtai, A., Elovici, Y.: Dynamic adversarial patch for evading object detection models. arXiv preprint arXiv:2010.13070  (2020)

\bibitem{hu2021naturalistic}
Hu, Y.C.T., Kung, B.H., Tan, D.S., Chen, J.C., Hua, K.L., Cheng, W.H.: Naturalistic physical adversarial patch for object detectors. In: Proceedings of the IEEE/CVF International Conference on Computer Vision. pp. 7848--7857 (2021)

\bibitem{hu2022adversarial}
Hu, Z., Huang, S., Zhu, X., Sun, F., Zhang, B., Hu, X.: Adversarial texture for fooling person detectors in the physical world. In: Proceedings of the IEEE/CVF conference on computer vision and pattern recognition. pp. 13307--13316 (2022)

\bibitem{huang2020universal}
Huang, L., Gao, C., Zhou, Y., Xie, C., Yuille, A.L., Zou, C., Liu, N.: Universal physical camouflage attacks on object detectors. In: Proceedings of the IEEE/CVF conference on computer vision and pattern recognition. pp. 720--729 (2020)

\bibitem{huang2023safari}
Huang, W., Zhao, X., Jin, G., Huang, X.: Safari: Versatile and efficient evaluations for robustness of interpretability. In: Proceedings of the IEEE/CVF International Conference on Computer Vision. pp. 1988--1998 (2023)

\bibitem{ilyas2018black}
Ilyas, A., Engstrom, L., Athalye, A., Lin, J.: Black-box adversarial attacks with limited queries and information. In: International conference on machine learning. pp. 2137--2146. PMLR (2018)

\bibitem{im2022adversarial}
Im~Choi, J., Tian, Q.: Adversarial attack and defense of yolo detectors in autonomous driving scenarios. In: 2022 IEEE Intelligent Vehicles Symposium (IV). pp. 1011--1017. IEEE (2022)

\bibitem{jia2022adv}
Jia, S., Yin, B., Yao, T., Ding, S., Shen, C., Yang, X., Ma, C.: Adv-attribute: Inconspicuous and transferable adversarial attack on face recognition. Advances in Neural Information Processing Systems  \textbf{35},  34136--34147 (2022)

\bibitem{jin2022enhancing}
Jin, G., Yi, X., Huang, W., Schewe, S., Huang, X.: Enhancing adversarial training with second-order statistics of weights. In: Proceedings of the IEEE/CVF conference on computer vision and pattern recognition. pp. 15273--15283 (2022)

\bibitem{jin2023randomized}
Jin, G., Yi, X., Wu, D., Mu, R., Huang, X.: Randomized adversarial training via taylor expansion. In: Proceedings of the IEEE/CVF Conference on Computer Vision and Pattern Recognition. pp. 16447--16457 (2023)

\bibitem{kahla2022label}
Kahla, M., Chen, S., Just, H.A., Jia, R.: Label-only model inversion attacks via boundary repulsion. In: Proceedings of the IEEE/CVF Conference on Computer Vision and Pattern Recognition. pp. 15045--15053 (2022)

\bibitem{kurakin2016adversarial}
Kurakin, A., Goodfellow, I., Bengio, S., et~al.: Adversarial examples in the physical world (2016)

\bibitem{lee2019physical}
Lee, M., Kolter, Z.: On physical adversarial patches for object detection. arXiv preprint arXiv:1906.11897  (2019)

\bibitem{lei2022using}
Lei, X., Cai, X., Lu, C., Jiang, Z., Gong, Z., Lu, L.: Using frequency attention to make adversarial patch powerful against person detector. IEEE Access  \textbf{11},  27217--27225 (2022)

\bibitem{li2019siamrpn++}
Li, B., Wu, W., Wang, Q., Zhang, F., Xing, J., Yan, J.: Siamrpn++: Evolution of siamese visual tracking with very deep networks. In: Proceedings of the IEEE/CVF conference on computer vision and pattern recognition. pp. 4282--4291 (2019)

\bibitem{li2018high}
Li, B., Yan, J., Wu, W., Zhu, Z., Hu, X.: High performance visual tracking with siamese region proposal network. In: Proceedings of the IEEE conference on computer vision and pattern recognition. pp. 8971--8980 (2018)

\bibitem{li2019adversarial}
Li, J., Schmidt, F., Kolter, Z.: Adversarial camera stickers: A physical camera-based attack on deep learning systems. In: International Conference on Machine Learning. pp. 3896--3904. PMLR (2019)

\bibitem{li2023physical}
Li, Y., Li, Y., Dai, X., Guo, S., Xiao, B.: Physical-world optical adversarial attacks on 3d face recognition. In: Proceedings of the IEEE/CVF Conference on Computer Vision and Pattern Recognition. pp. 24699--24708 (2023)

\bibitem{liang2022parallel}
Liang, S., Wu, B., Fan, Y., Wei, X., Cao, X.: Parallel rectangle flip attack: A query-based black-box attack against object detection. arXiv preprint arXiv:2201.08970  (2022)

\bibitem{lin2014microsoft}
Lin, T.Y., Maire, M., Belongie, S., Hays, J., Perona, P., Ramanan, D., Doll{\'a}r, P., Zitnick, C.L.: Microsoft coco: Common objects in context. In: Computer Vision--ECCV 2014: 13th European Conference, Zurich, Switzerland, September 6-12, 2014, Proceedings, Part V 13. pp. 740--755. Springer (2014)

\bibitem{liu2019perceptual}
Liu, A., Liu, X., Fan, J., Ma, Y., Zhang, A., Xie, H., Tao, D.: Perceptual-sensitive gan for generating adversarial patches. In: Proceedings of the AAAI conference on artificial intelligence. vol.~33, pp. 1028--1035 (2019)

\bibitem{liu2023slowlidar}
Liu, H., Wu, Y., Yu, Z., Vorobeychik, Y., Zhang, N.: Slowlidar: Increasing the latency of lidar-based detection using adversarial examples. In: Proceedings of the IEEE/CVF Conference on Computer Vision and Pattern Recognition. pp. 5146--5155 (2023)

\bibitem{liu2018dpatch}
Liu, X., Yang, H., Liu, Z., Song, L., Li, H., Chen, Y.: Dpatch: An adversarial patch attack on object detectors. arXiv preprint arXiv:1806.02299  (2018)

\bibitem{lovisotto2021slap}
Lovisotto, G., Turner, H., Sluganovic, I., Strohmeier, M., Martinovic, I.: $\{$SLAP$\}$: Improving physical adversarial examples with $\{$Short-Lived$\}$ adversarial perturbations. In: 30th USENIX Security Symposium (USENIX Security 21). pp. 1865--1882 (2021)

\bibitem{madry2017towards}
Madry, A., Makelov, A., Schmidt, L., Tsipras, D., Vladu, A.: Towards deep learning models resistant to adversarial attacks. arXiv preprint arXiv:1706.06083  (2017)

\bibitem{mahendran2015understanding}
Mahendran, A., Vedaldi, A.: Understanding deep image representations by inverting them. In: Proceedings of the IEEE conference on computer vision and pattern recognition. pp. 5188--5196 (2015)

\bibitem{moosavi2016deepfool}
Moosavi-Dezfooli, S.M., Fawzi, A., Frossard, P.: Deepfool: a simple and accurate method to fool deep neural networks. In: Proceedings of the IEEE conference on computer vision and pattern recognition. pp. 2574--2582 (2016)

\bibitem{pomponi2022pixle}
Pomponi, J., Scardapane, S., Uncini, A.: Pixle: a fast and effective black-box attack based on rearranging pixels. In: 2022 International Joint Conference on Neural Networks (IJCNN). pp.~1--7. IEEE (2022)

\bibitem{schroff2015facenet}
Schroff, F., Kalenichenko, D., Philbin, J.: Facenet: A unified embedding for face recognition and clustering. In: Proceedings of the IEEE conference on computer vision and pattern recognition. pp. 815--823 (2015)

\bibitem{shapira2023phantom}
Shapira, A., Zolfi, A., Demetrio, L., Biggio, B., Shabtai, A.: Phantom sponges: Exploiting non-maximum suppression to attack deep object detectors. In: Proceedings of the IEEE/CVF Winter Conference on Applications of Computer Vision. pp. 4571--4580 (2023)

\bibitem{sharif2016accessorize}
Sharif, M., Bhagavatula, S., Bauer, L., Reiter, M.K.: Accessorize to a crime: Real and stealthy attacks on state-of-the-art face recognition. In: Proceedings of the 2016 acm sigsac conference on computer and communications security. pp. 1528--1540 (2016)

\bibitem{shi2023reinforcement}
Shi, Z., Yang, W., Xu, Z., Yu, Z., Huang, L.: Reinforcement learning-based adversarial attacks on object detectors using reward shaping. In: Proceedings of the 31st ACM International Conference on Multimedia. pp. 8424--8432 (2023)

\bibitem{szegedy2013intriguing}
Szegedy, C., Zaremba, W., Sutskever, I., Bruna, J., Erhan, D., Goodfellow, I., Fergus, R.: Intriguing properties of neural networks. arXiv preprint arXiv:1312.6199  (2013)

\bibitem{thys2019fooling}
Thys, S., Van~Ranst, W., Goedem{\'e}, T.: Fooling automated surveillance cameras: adversarial patches to attack person detection. In: Proceedings of the IEEE/CVF conference on computer vision and pattern recognition workshops. pp.~0--0 (2019)

\bibitem{tu2020physically}
Tu, J., Ren, M., Manivasagam, S., Liang, M., Yang, B., Du, R., Cheng, F., Urtasun, R.: Physically realizable adversarial examples for lidar object detection. In: Proceedings of the IEEE/CVF Conference on Computer Vision and Pattern Recognition. pp. 13716--13725 (2020)

\bibitem{wang2022fca}
Wang, D., Jiang, T., Sun, J., Zhou, W., Gong, Z., Zhang, X., Yao, W., Chen, X.: Fca: Learning a 3d full-coverage vehicle camouflage for multi-view physical adversarial attack. In: Proceedings of the AAAI conference on artificial intelligence. vol.~36, pp. 2414--2422 (2022)

\bibitem{wang2023rfla}
Wang, D., Yao, W., Jiang, T., Li, C., Chen, X.: Rfla: A stealthy reflected light adversarial attack in the physical world. In: Proceedings of the IEEE/CVF International Conference on Computer Vision. pp. 4455--4465 (2023)

\bibitem{wang2018cosface}
Wang, H., Wang, Y., Zhou, Z., Ji, X., Gong, D., Zhou, J., Li, Z., Liu, W.: Cosface: Large margin cosine loss for deep face recognition. In: Proceedings of the IEEE conference on computer vision and pattern recognition. pp. 5265--5274 (2018)

\bibitem{wang2021dual}
Wang, J., Liu, A., Yin, Z., Liu, S., Tang, S., Liu, X.: Dual attention suppression attack: Generate adversarial camouflage in physical world. In: Proceedings of the IEEE/CVF Conference on Computer Vision and Pattern Recognition. pp. 8565--8574 (2021)

\bibitem{wang2019fast}
Wang, Q., Zhang, L., Bertinetto, L., Hu, W., Torr, P.H.: Fast online object tracking and segmentation: A unifying approach. In: Proceedings of the IEEE/CVF conference on Computer Vision and Pattern Recognition. pp. 1328--1338 (2019)

\bibitem{wang2021enhancing}
Wang, X., He, K.: Enhancing the transferability of adversarial attacks through variance tuning. In: Proceedings of the IEEE/CVF Conference on Computer Vision and Pattern Recognition. pp. 1924--1933 (2021)

\bibitem{wu2020skip}
Wu, D., Wang, Y., Xia, S.T., Bailey, J., Ma, X.: Skip connections matter: On the transferability of adversarial examples generated with resnets. arXiv preprint arXiv:2002.05990  (2020)

\bibitem{wu2020boosting}
Wu, W., Su, Y., Chen, X., Zhao, S., King, I., Lyu, M.R., Tai, Y.W.: Boosting the transferability of adversarial samples via attention. In: Proceedings of the IEEE/CVF Conference on Computer Vision and Pattern Recognition. pp. 1161--1170 (2020)

\bibitem{xie2019improving}
Xie, C., Zhang, Z., Zhou, Y., Bai, S., Wang, J., Ren, Z., Yuille, A.L.: Improving transferability of adversarial examples with input diversity. In: Proceedings of the IEEE/CVF conference on computer vision and pattern recognition. pp. 2730--2739 (2019)

\bibitem{xu2020adversarial}
Xu, K., Zhang, G., Liu, S., Fan, Q., Sun, M., Chen, H., Chen, P.Y., Wang, Y., Lin, X.: Adversarial t-shirt! evading person detectors in a physical world. In: Computer Vision--ECCV 2020: 16th European Conference, Glasgow, UK, August 23--28, 2020, Proceedings, Part V 16. pp. 665--681. Springer (2020)

\bibitem{zhang2022improving}
Zhang, J., Wu, W., Huang, J.t., Huang, Y., Wang, W., Su, Y., Lyu, M.R.: Improving adversarial transferability via neuron attribution-based attacks. In: Proceedings of the IEEE/CVF Conference on Computer Vision and Pattern Recognition. pp. 14993--15002 (2022)

\bibitem{zhang2022towards}
Zhang, J., Li, B., Xu, J., Wu, S., Ding, S., Zhang, L., Wu, C.: Towards efficient data free black-box adversarial attack. In: Proceedings of the IEEE/CVF Conference on Computer Vision and Pattern Recognition. pp. 15115--15125 (2022)

\bibitem{zhang2023trajpac}
Zhang, L., Xu, N., Yang, P., Jin, G., Huang, C.C., Zhang, L.: Trajpac: Towards robustness verification of pedestrian trajectory prediction models. In: Proceedings of the IEEE/CVF International Conference on Computer Vision. pp. 8327--8339 (2023)

\bibitem{zhao2019seeing}
Zhao, Y., Zhu, H., Liang, R., Shen, Q., Zhang, S., Chen, K.: Seeing isn't believing: Towards more robust adversarial attack against real world object detectors. In: Proceedings of the 2019 ACM SIGSAC conference on computer and communications security. pp. 1989--2004 (2019)

\bibitem{zhong2022shadows}
Zhong, Y., Liu, X., Zhai, D., Jiang, J., Ji, X.: Shadows can be dangerous: Stealthy and effective physical-world adversarial attack by natural phenomenon. In: Proceedings of the IEEE/CVF Conference on Computer Vision and Pattern Recognition. pp. 15345--15354 (2022)

\bibitem{zhu2023efficient}
Zhu, L., Wang, T., Li, J., Zhang, Z., Shen, J., Wang, X.: Efficient query-based black-box attack against cross-modal hashing retrieval. ACM Transactions on Information Systems  \textbf{41}(3),  1--25 (2023)

\bibitem{zhu2021fooling}
Zhu, X., Li, X., Li, J., Wang, Z., Hu, X.: Fooling thermal infrared pedestrian detectors in real world using small bulbs. In: Proceedings of the AAAI Conference on Artificial Intelligence. vol.~35, pp. 3616--3624 (2021)

\bibitem{zolfi2021translucent}
Zolfi, A., Kravchik, M., Elovici, Y., Shabtai, A.: The translucent patch: A physical and universal attack on object detectors. In: Proceedings of the IEEE/CVF Conference on Computer Vision and Pattern Recognition. pp. 15232--15241 (2021)

\end{thebibliography}

\newpage
\setcounter{page}{1}
\end{document}